\documentclass{article}
\usepackage{custom_tex}
\usepackage{placeins}

\newcommand{\pmsd}[2]{#1{\scriptsize$\pm$#2}}

\title{Transport along Local Annealing Paths: \\ Wasserstein Geometry and Multimodal Sampling}

\author[1]{Han Chen}
\author[1]{Sifan Liu}
\affil[1]{Department of Statistical Science, Duke University}

\begin{document}
\maketitle

\begin{abstract}
    Variational approaches to learning transport maps for multimodal sampling can suffer from mode collapse. One way to mitigate this problem is to learn transport along a prescribed path of intermediate distributions defined by geometric annealing. We show that annealing can take a substantial detour in Wasserstein space: when mass moves between modes, the path can be much longer than the distance between its endpoints. We further link path geometry to transport learning, showing that long annealing paths can slow population optimization, while sufficiently regular paths permit rapid convergence. Motivated by these findings, we propose componentwise annealed transport (CAT) to reduce the need for mass transfer between modes. Starting from a mode-informed mixture reference, CAT constructs an exact mixture decomposition of the target and learns separate transport maps along componentwise annealing paths. We estimate the component weights separately and show that these estimates remain consistent even when the learned flows are inexact. We also develop adaptive annealing schedules guided by transport effort. Experiments on multimodal synthetic distributions and Bayesian inference problems demonstrate improvements in transport learning and sampling efficiency.
\end{abstract}

\section{Introduction}
Sampling from multimodal distributions can be challenging when regions of high probability are well-separated. This separation can hinder exploration and make it difficult to obtain representative samples.
Annealing and tempering offer a classical approach: they introduce intermediate distributions between a tractable reference and the target. At higher temperatures, these distributions become flatter, facilitating movement between modes. This strategy underlies methods such as annealed importance sampling~\citep[AIS;][]{neal2001annealed}, tempered transitions~\citep{neal1996sampling}, sequential Monte Carlo~\citep[SMC;][]{del2006sequential}, and simulated and parallel tempering~\citep{swendsen1986replica,marinari1992simulated,geyer1995annealing}.

Classical annealing methods propagate samples through intermediate distributions via stochastic transitions, often combined with importance weighting, resampling, or Metropolis adjustment. Learning-based methods instead construct reusable transport maps for this propagation. Once learned, a map can efficiently generate independent samples from an approximation to the target, amortizing its training cost, whereas MCMC requires further simulation to obtain new samples. Transport maps can be parameterized using Knothe--Rosenblatt maps~\citep{el2012bayesian}, normalizing flows~\citep{papamakarios2021normalizing}, or continuous normalizing flows~\citep{chen2018neural}, or constructed sequentially via iterative Gaussianization~\citep{chen2025rotated}. Yet learning a map directly for a multimodal target remains challenging: variational training that minimizes the reverse Kullback--Leibler (KL) divergence can favor approximations concentrated on a subset of the target modes, a phenomenon known as mode collapse~\citep{soletskyi2025theoretical}.

Several methods combine annealing with transport maps, but use intermediate distributions for different purposes. Within SMC, annealed flow transport \citep[AFT;][]{arbel2021annealed} and continual repeated annealed flow transport \citep[CRAFT;][]{matthews2022continual} learn maps between successive annealed targets alongside importance weighting, resampling, and Markov transitions. In parallel tempering, learned transports can instead assist exchanges between neighboring temperatures~\citep{zhang2026accelerated}. Annealing can also support flow training: Flow AIS bootstrap~\citep[FAB;][]{midgley2023flow} uses AIS-generated samples to train a flow with a mass-covering $\chi^2$-divergence, whereas TemperFlow~\citep{qiu2024efficient} fits a transport map successively to tempered targets, using the parameters learned at one temperature to warm-start training at the next. Annealing Flow learns a sequence of continuous normalizing-flow blocks targeting successive annealed distributions, using a reverse-KL objective regularized by transport energy~\citep{wu2025annealing}. A theoretical study of variational inference for Gaussian mixtures examines how the annealing schedule affects mode collapse during reverse-KL optimization~\citep{fogliani2026annealing,wang2026mitigating}. Gibbs flow constructs approximate continuous-time transports along annealing paths using full conditional distributions and incorporates them into SMC samplers~\citep{heng2021gibbs}. Other continuous-time methods learn velocity fields along prescribed or learned density paths using continuity-equation residuals, sometimes with importance weights to correct imperfect transport~\citep{mate2023learning,tian2024liouville,albergo2025nets}. Kernel methods also construct transport along prescribed annealing paths by approximating velocity fields in reproducing kernel Hilbert spaces~\citep{maurais2024sampling,nusken2024stein}.

However, learning transport along an annealing path can be difficult when the relative masses of modes change with temperature. A velocity field that realizes the path must then move probability between modes. If those modes are separated by regions of very low probability, this transfer can require large velocities, making the flow difficult to learn. Changes in modal probabilities also contribute to slow mixing in parallel and simulated tempering~\citep{woodard2009torpid}, while difficulties with inter-mode transport have been observed in learning density interpolations~\citep{mate2023learning}.

One solution is to modify the annealing path to preserve modal weights. Weight-preserving annealing has been proposed for simulated and parallel tempering~\citep{tawn2020weight}. It uses mode locations and local curvature information to define a hard partition of the sample space that approximately preserves modal weights. However, the resulting density may be discontinuous at partition boundaries. Another approach is to construct alternative probability paths between the reference and target distributions. For example, diffusion paths progressively add noise to target samples, preserving mixture weights while smoothing the components. However, the intermediate densities and scores are generally unavailable in closed form. Stochastic optimal control formulations provide a way to learn the drift or score functions needed for sampling~\citep{zhang2022path,vargas2023denoising}. Practical training generally lacks global convergence guarantees, and errors in the learned scores and time discretization can bias samples. Metropolis correction can remove this sampling bias, although its acceptance probability remains sensitive to score approximation errors~\citep{chen2026markov}.

We propose a new probability path based on local annealing. Starting from a mode-informed mixture reference, we decompose the target into overlapping mixture components and anneal each component separately, keeping the reference mixture weights fixed along the path. This separates the learning of local transport from the estimation of component weights, avoiding the need to move mass between components merely to adjust their relative probabilities. When the reference components capture the target's local structure, the resulting componentwise paths can avoid the transport bottlenecks of global annealing. Unlike hard-partition constructions~\citep{tawn2020weight}, this soft decomposition does not introduce discontinuities at partition boundaries. We learn a transport map for each component and then calibrate the mixture weights to approximate the target.

We characterize the benefits of componentwise annealing through a study of Wasserstein geometry and optimization dynamics. Poincar\'e inequalities are widely used to analyze the convergence of Langevin dynamics~\citep{bakry2014analysis,chewi2022analysis}; our analysis identifies their complementary role in transport learning. A uniform Poincar\'e inequality along the path both controls the Wasserstein length of the path and supports fast population optimization. This connection provides further motivation for constructing regular local paths to facilitate efficient transport learning.
The main contributions are summarized as follows.
\begin{itemize}
    \item \textbf{A geometric characterization of annealing paths.}
    We study annealing paths in Wasserstein space and quantify how moving probability across low-density regions can make a path substantially longer than the Wasserstein distance between its endpoints. In contrast, under a uniform Poincar\'e inequality, the path length can be bounded in terms of the Poincar\'e constant and a divergence between only the two endpoints.

    \item \textbf{A link between path geometry and optimization dynamics.}
    We analyze the population gradient flow of the continuity-equation residual loss in the space of velocity fields. We show that a long annealing path imposes a lower bound on accumulated optimization loss and can lead to slow convergence. Conversely, a uniform Poincar\'e inequality implies a Polyak--\L{}ojasiewicz inequality and exponential loss decay.

    \item \textbf{A practical algorithm with consistent weight estimation and adaptive schedules.}
    We develop a practical algorithm that separates component transport from weight estimation, allowing the component flows to be trained independently and in parallel. Under a support condition, the component weight estimator is consistent even when the learned flows are inexact. We also develop component-specific adaptive schedules guided by a bound on the transport effort between successive targets.
\end{itemize}

The remainder of the paper is organized as follows.
Section~\ref{sec:annealing_geometry} studies the Wasserstein geometry of annealing paths.
Section~\ref{sec:local_annealing} constructs componentwise annealing paths from a mixture reference and bounds their Wasserstein length.
Section~\ref{sec:velocity-optimization} formulates the velocity-field optimization problem and relates path geometry to its population gradient-flow dynamics.
Section~\ref{sec:algorithm} presents the main algorithm, including weight calibration and adaptive annealing schedules.
Section~\ref{sec:experiments} reports numerical experiments on synthetic and Bayesian sampling problems.
Section~\ref{sec:discussion} concludes with a discussion of future directions.
Proofs and experimental details are deferred to the appendices.

\section{Geometry of annealing paths}
\label{sec:annealing_geometry}

Learning transport maps along a prescribed path of intermediate distributions depends on the entire path, not just its endpoints. We therefore study the path's Wasserstein geometry and quantify the cost of realizing it.

\subsection{Wasserstein geometry of probability paths}
\label{subsec:wasserstein_geometry}

Let \(\mathcal P_2(\mathbb R^d)\) be the set of Borel probability measures on \(\mathbb R^d\) with finite second moments. A probability path $(\rho_\beta)_{\beta\in[0,1]}$ is a family of measures in this set indexed by \(\beta\in[0,1]\). A velocity field $(v_\beta)_{\beta\in[0,1]}$ generates the path if it satisfies the continuity equation
\begin{equation}\label{eq:CE}
    \partial_\beta \rho_\beta + \nabla\cdot(\rho_\beta v_\beta)=0
\end{equation}
in the distributional sense. Under the integrability and Lipschitz conditions of \citet[Propositions 8.1.7]{ambrosio2005gradient}, the associated particle ODE represents this path: if $X_0\sim \rho_0$ and $\dot X_\beta=v_\beta(X_\beta)$, then $X_\beta\sim \rho_\beta$ for all $\beta\in[0,1]$. 
To see this, apply the chain rule to any $f\in C_c^\infty(\R^d)$ to obtain
\begin{align*}
    \partial_\beta \int f \rd \rho_\beta = \partial_\beta \EE{f(X_\beta)} = \EE{\nabla f(X_\beta) \cdot v_\beta(X_\beta)} = \int \nabla f\cdot v_\beta\,\rd\rho_\beta,
\end{align*}
which is the weak form of the continuity equation.
Here and below, $\nabla$ denotes the gradient with respect to $x$; integrals written with $\rd\rho_\beta$ are taken against the measure $\rho_\beta$.

For \(\mu,\nu\in\mathcal P_2(\mathbb R^d)\), the \(2\)-Wasserstein distance is defined as
\begin{equation*}
    W_2(\mu,\nu):=
    \inf_{\substack{(v, \rho):\ \partial_\beta \rho_\beta + \nabla\cdot(\rho_\beta v_\beta)=0,\\ \rho_0=\mu,\ \rho_1=\nu }}\;\;
    \int_0^1
    \left\{
    \int_{\mathbb R^d}\|v_\beta(x)\|^2\,\rho_\beta(\rd x)
    \right\}^{1/2} \rd\beta.
\end{equation*}
The infimum is over Borel velocity fields \(v_\beta\) and weakly continuous paths \(\rho_\beta\in \calP_2(\R^d)\) satisfying the continuity equation and the endpoint constraints. Thus, $W_2(\mu,\nu)$ is the minimum transport length among paths joining \(\mu\) and \(\nu\). The metric space \((\mathcal P_2(\mathbb R^d),W_2)\) is called the \(2\)-Wasserstein space.

We next define the length of a probability path $(\rho_\beta)_{\beta\in[0,1]}$ in Wasserstein space.
A curve \(\rho=(\rho_\beta)_{\beta\in[0,1]}\) in \((\mathcal P_2(\mathbb R^d),W_2)\) is absolutely continuous if there exists a nonnegative \(g\in L^1(0,1)\) such that
$W_2(\rho_s,\rho_t)\leq\int_s^t g(\beta)\,\rd\beta$ for any $0\leq s<t\leq1$.
Its \emph{Wasserstein speed} is the metric derivative
\begin{equation} \label{eq:wasserstein_speed}
    |\dot\rho|(\beta)
    :=
    \lim_{\varepsilon\to0}
    \frac{W_2(\rho_{\beta+\varepsilon},\rho_\beta)}{|\varepsilon|}
    =\inf_{v_\beta:\,\partial_\beta\rho_\beta+\nabla\cdot(\rho_\beta v_\beta)=0}\;
    \left(\int_{\mathbb{R}^d}
    \|v_\beta(x)\|^2
    \,\rho_\beta(\rd x)\right)^{1/2},
\end{equation}
for almost every $\beta\in[0,1]$.
By \citet[Theorem~8.3.1]{ambrosio2005gradient}, the infimum is attained by a unique minimum-norm velocity in the \(L^2(\rho_\beta)\)-closure of gradients of smooth compactly supported functions. 
The \emph{Wasserstein length} of an absolutely continuous path \(\rho\) is its total distance traveled \citep[Section~1.1]{ambrosio2005gradient}:
\[
    \begin{aligned}
    \mathcal{L}(\rho)
    :=
    \int_0^1
    |\dot\rho|(\beta)
    \,\rd\beta
    &=\inf_{v:\,\partial_\beta\rho_\beta+\nabla\cdot(\rho_\beta v_\beta)=0}
    \int_0^1
    \left\{
    \int_{\mathbb R^d}\|v_\beta(x)\|^2\,\rho_\beta(\rd x)
    \right\}^{1/2}\,\rd\beta.
    \end{aligned}
\]
The second equality follows from~\eqref{eq:wasserstein_speed} and the existence of a minimum-norm velocity field.
Thus, $\mathcal L(\rho)\geq W_2(\rho_0,\rho_1)$: the Wasserstein distance is the minimum length between the endpoints, whereas $\mathcal L(\rho)$ is the length of a particular path. Equality holds when the curve $\rho$ is precisely a $W_2$ geodesic joining $\rho_0$ and $\rho_1$.

Path geometry characterizes the transport required to realize a prescribed probability path. In our sampling task, a shorter path requires less cumulative transport, potentially making its velocity field easier to learn. Section~\ref{sec:velocity-optimization} relates Wasserstein length to velocity-field optimization, making this connection precise. Path geometry also informs the analysis of stochastic samplers: \citet{guo2025provable} derive convergence bounds for annealed Langevin Monte Carlo that depend on the Wasserstein action of the annealing path.

\subsection{Wasserstein length of annealing paths}
\label{sec:global_transport_cost}

Let $p,b\in\calP_2(\R^d)$ be the target and reference distributions, respectively, and assume that both have strictly positive $C^1(\R^d)$ densities with respect to the Lebesgue measure. We use the same symbols for these distributions and their densities. A natural path connecting them is geometric annealing, $p_\beta\propto p^\beta b^{1-\beta}$, which moves from $b$ to $p$ as $\beta$ ranges from 0 to 1. This path can be written as
\begin{equation}
\label{eq:global_annealing}
    p_\beta(x)=\frac{b(x)e^{\beta h(x)}}{Z_\beta},\;\; \beta\in[0,1],
    \qquad
    h = \log p - \log b,
\end{equation}
where $Z_\beta$ is the normalizing constant. We also assume that $(p_\beta)_{\beta\in[0,1]}$ is an absolutely continuous curve in \((\mathcal P_2(\mathbb R^d),W_2)\). 

For a multimodal target, annealing can change the relative masses of the modes. A velocity field that realizes such a path must transport mass between modes, potentially resulting in a large Wasserstein length. We first quantify this effect through a lower bound based on a smooth observable.

\begin{lemma}[Wasserstein length lower bound]
\label{lem:lower_bound}
Let $\rho=(\rho_\beta)_{\beta\in[0,1]}$ be an absolutely continuous curve in $(\mathcal P_2(\mathbb R^d),W_2)$, and let $\chi:\mathbb R^d\to\mathbb R$ be continuously differentiable with bounded gradient. Define
$m_\chi(\beta):=\E_{\rho_\beta}[\chi]$ and
$s_\chi^2(\beta):=\E_{\rho_\beta}[\|\nabla\chi\|_2^2]$.
If $s_\chi(\beta)\leq\varepsilon$ for some $\varepsilon>0$ and almost every
$\beta\in[\beta_0,\beta_1]\subset [0,1]$, then
\begin{equation*}
    \mathcal L(\rho)
    \geq
    \frac{|m_\chi(\beta_1)-m_\chi(\beta_0)|}{\varepsilon}.
\end{equation*}
\end{lemma}
The proof is given in Appendix~\ref{prf:lower_bound}.
Suppose the target $p$ has several modes separated by regions of low probability. Choose a smooth function \(\chi:\mathbb R^d\to[0,1]\) that is approximately one near one mode and zero elsewhere, so that \(m_\chi(\beta)\) approximates that mode's mass. If \(\|\nabla\chi\|\leq M\), \(\nabla\chi\) is supported in a separating region \(B\), and $p_\beta(B)\leq\delta$ for $\beta\in[\beta_0,\beta_1]$, then Lemma~\ref{lem:lower_bound} gives
\[
\mathcal L(p)\geq \frac{|m_\chi(\beta_1)-m_\chi(\beta_0)|}{M\sqrt\delta}.
\]
Here, the numerator measures the change in modal mass, while the factor $\sqrt\delta$ captures the low probability of the separating region. Thus, a substantial change in the masses of modes separated by low-probability regions necessarily forces a long Wasserstein path. The following example makes this mechanism explicit.

\begin{corollary}[Exponential length of Gaussian-mixture annealing]
\label{coro:gaussian-annealing-length}
Let \(f_-\) and \(f_+\) be the densities of the two $d$-dimensional Gaussians \(\mathcal N(-\mathbf 1_d,I_d)\) and \(\mathcal N(\mathbf 1_d,I_d)\), respectively. Assume $d\geq1$ and let $b=f_-$, $p=\tfrac12 f_-+\tfrac12 f_+$. Then the Wasserstein length of the geometric annealing path from \(b\) to \(p\) has the lower bound
\[
    \mathcal L(p)\geq\frac{\sqrt d}{8}e^{d/16}.
\]
\end{corollary}

The proof is given in Appendix~\ref{prf:gaussian-annealing-length}.
In this example, \(W_2(b,p)\leq\sqrt{2d}\), yet the Wasserstein length of the geometric annealing path grows exponentially with \(d\). The gap shows that this prescribed path requires far more cumulative transport than the shortest path between its endpoints because it shifts mass between the modes while the separating region remains very unlikely throughout the path. Therefore, annealing paths are highly inefficient in terms of Wasserstein transport.

\section{Componentwise annealing}
\label{sec:local_annealing}

The preceding example shows that geometric annealing can have a large Wasserstein length when it shifts mass between separated modes. This motivates a more localized form of annealing that avoids transporting mass between those modes.

\subsection{From a mixture base to componentwise annealing paths}
\label{sec:mixture_decomposition}

We distinguish three tasks in multimodal sampling: locating the modes, learning transport within each mode, and estimating their relative weights. Separating within-mode transport from the estimation of relative weights parallels multimodal MCMC strategies that distinguish within-mode exploration from global moves~\citep{pompe2020framework,tawn2021annealed,tan2025accelerate}.

In this section, we assume that some information about the modes is available. For example, a preliminary search can locate them, and local Laplace approximations around these locations can then provide initial component distributions~\citep{tawn2021annealed,tan2025accelerate}. Details of this construction are given in Appendix~\ref{sec:experiment-base-training}. These components define the mixture base distribution
\[
    b(x)=\sum_{j=1}^J a_jb_j(x),
\]
where each $b_j\in C^1(\R^d)$ is a positive density and the weights satisfy $a_j>0$ and $\sum_{j=1}^J a_j=1$.

The mixture base induces an exact decomposition of the target $p$. For each $x\in\R^d$, define the soft assignment weight
\begin{equation*}
    r_j(x):=\frac{a_jb_j(x)}{b(x)},
    \qquad j\in[J]=\{ 1,\ldots,J\}.
\end{equation*}
The corresponding target component is
\begin{align*}
     p_j(x):=\frac{r_j(x)p(x)}{w_j},\qquad w_j:=\int r_j(x)p(x)\,\rd x = \E_p[r_j].
\end{align*}
We assume $w_j>0$ for every $j$, which holds when $b_j$ is positive on a set of positive $p$-probability. Since the assignment weights sum to one, the target is the mixture
\begin{equation*}
    p(x)=\sum_{j=1}^J r_j(x)p(x)= \sum_{j=1}^Jw_jp_j(x).
\end{equation*}
This identity holds for any choice of $b$ as long as $w_j>0$ for every $j$.

We now connect each base component $b_j$ to its corresponding target component $p_j$ along a separate geometric annealing path. For $j\in[J]$, let $p_\beta^j(x)\propto b_j(x)^{1-\beta}p_j(x)^\beta$, so that $p_0^j=b_j$ and $p_1^j=p_j$. Using the common log density ratio $h=\log(p/b)$ introduced above, we can also write this path in exponential-tilting form:
\begin{equation} 
    \label{eq:local-annealing-path}
    p_\beta^j(x)=\frac{b_j(x)e^{\beta h(x)}}{Z_\beta^j},
    \qquad
    Z_\beta^j:=\int_{\mathbb R^d}b_j(x)e^{\beta h(x)}\,\rd x.
\end{equation}
We then take a weighted average of these paths using the fixed reference weights $a_j$:
\begin{align}\label{eq:def-componentwise-annealing-path}
    \bar p_\beta(x):=\sum_{j=1}^J a_j p_\beta^j(x).
\end{align}
We call $(\bar p_\beta)_{\beta\in[0,1]}$ the \emph{componentwise annealing} path. It starts at $\bar p_0=b$ and ends at
\[
\bar p_1=\sum_{j=1}^J a_j p_j =: p^{(a)}.
\]
The endpoint equals the target $p$ when $a_j=w_j$ for every $j$, but may differ otherwise. Because the weights $a_j$ remain fixed along the path, this construction does not require transporting mass between components. Matching the target instead requires separate weight calibration, as discussed in Section~\ref{sec:weight-calibration}.

For comparison, the global annealing path~\eqref{eq:global_annealing} from $b$ to $p$ can be written using the same component paths:
\[
    p_\beta=\sum_{j=1}^J\lambda_j(\beta)p_\beta^j,
    \qquad
    \lambda_j(\beta):=
    \frac{a_jZ_\beta^j}{\sum_{\ell=1}^Ja_\ell Z_\beta^\ell}.
\]
Here \(Z_\beta^j\) is the normalizing constant defined in~\eqref{eq:local-annealing-path}. Both paths start at $b$ and use the same component distributions at each $\beta$; only their mixture weights differ. In global annealing, $\lambda_j(\beta)$ can vary with $\beta$, whereas componentwise annealing fixes the weights at $a_j$.
As shown above, changing mixture weights can make a path long when the components represent well-separated modes. We next bound the length of the fixed-weight componentwise path in terms of the regularity of its component paths and a divergence between each pair of endpoints $b_j$ and $p_j$.

\subsection{Wasserstein length of componentwise annealing}
\label{sec:componentwise-wasserstein-length}

Along the componentwise annealing path~\eqref{eq:def-componentwise-annealing-path}, the mixture weights remain fixed, so each component can change without altering its share of the mixture. The path can therefore be short even when components are far apart, provided each $b_j$ is close to its corresponding $p_j$ and the intermediate component distributions remain regular. We formalize this regularity through a uniform Poincar\'e inequality.

\begin{assumption}[Componentwise Poincar\'e inequality]
\label{assump:poincare}
For each \(j\in[J]\), the intermediate distributions \(p_\beta^j\) satisfy a Poincar\'e inequality with constant $C_{\mathrm{P}, j}$ uniform in $\beta$. Specifically, for any smooth function \(f\),
\begin{equation}
    \label{eq:component_uniform_poincare}
    \operatorname{Var}_{p_\beta^j}(f)
    \leq C_{\mathrm{P}, j}\, \E_{p_\beta^j}[\|\nabla f\|^2],
    \qquad \beta\in[0,1].
\end{equation}
\end{assumption}
Strong log-concavity is one sufficient condition: if both $b_j$ and $p_j$ are strongly log-concave with constant $1/C$, every intermediate $p_\beta^j$ is also strongly log-concave with that constant, so $C_{\mathrm P,j}\leq C$.

\begin{theorem}[Wasserstein length of componentwise annealing]
\label{thm:componentwise-annealing-length}
Under Assumption~\ref{assump:poincare} and the standing regularity conditions, the Wasserstein length of the componentwise annealing path \eqref{eq:def-componentwise-annealing-path} satisfies
\begin{equation*}
    \mathcal L(\bar p)^2
    \,\leq\, \sum_{j=1}^J a_j\, C_{\mathrm P,j}
    \,\Bigl\{\mathrm{KL}(p_j\Vert b_j)+\mathrm{KL}(b_j\Vert p_j)\Bigr\}.
\end{equation*}
\end{theorem}

The bound follows from a simple construction. If $v_\beta^j$ realizes the component path $p_\beta^j$, then the weighted velocity
\begin{align}\label{eq:def-weighted-velocity}
    \bar v_\beta(x) := \sum_{j=1}^J \frac{ a_j p_\beta^j(x) }{\bar p_\beta(x)} v_\beta^j(x)
\end{align}
realizes the componentwise path $\bar p_\beta$. Thus, the component velocities can be learned separately and then combined. Convexity bounds the resulting Wasserstein speed; the proof is given in Appendix~\ref{prf:componentwise-annealing-length}.
\citet{chemseddine2025neural} use a Poincar\'e inequality to bound the transport energy of a probability path by its Fisher--Rao action multiplied by the Poincar\'e constant. Our bound applies this principle componentwise and expresses the resulting bound in terms of the symmetrized KL divergences between component endpoints.

The upper bound depends on the mismatch within each pair $(b_j,p_j)$ and on the regularity of the path between them. In particular, if the Poincar\'e constants remain bounded and the endpoint divergences are small, the componentwise path is short. Crucially, the bound has no explicit dependence on the distances between different components or the relative weights or scales of different components.
In this sense, componentwise annealing is a local counterpart to global annealing: it evolves each component along its own path while keeping the mixture weights fixed, avoiding the inter-mode mass transfer that can lead to long Wasserstein paths.

\section{Velocity optimization and gradient-flow dynamics}
\label{sec:velocity-optimization}

We now formulate the optimization problem for learning a velocity field that realizes the annealing path. We then show how the resulting population gradient flow connects to the Wasserstein geometry of the path studied in the preceding sections.

\subsection{Continuity-equation residual and population loss}
\label{sec:population-residual}

As discussed in Section~\ref{sec:componentwise-wasserstein-length}, velocity fields that realize the component paths can be combined, as in Equation~\eqref{eq:def-weighted-velocity}, to obtain a velocity field for the componentwise annealing path. Therefore, it suffices to analyze one geometric annealing path. We use the global annealing path~\eqref{eq:global_annealing} for notation; the same formulation applies to each component path separately.

To realize the annealing path, a velocity field must satisfy the continuity equation~\eqref{eq:CE}. Since $\partial_\beta \log p_\beta = h - \partial_\beta \log Z_\beta$, this equation takes the following logarithmic form:
\[
h - \partial_\beta \log Z_\beta + \nabla \log p_\beta \cdot v_\beta + \nabla\cdot v_\beta = 0.
\]
The derivative $\partial_\beta\log Z_\beta$ is unknown, so we represent it by a scalar function $c_\beta$ and optimize $c$ jointly with $v$. For this optimization, we use the real Hilbert space
\[
 \mathcal H
 :=L^2\!\left([0,1]\times\mathbb R^d,
              \rd \beta p_\beta(\rd x);\mathbb R^d\right)
   \times L^2(0,1).
\]
An element $(v,c)\in\mathcal H$ consists of a measurable velocity field $v_\beta:\R^d\to \R^d$ and a scalar $c_\beta\in\mathbb R$, both indexed by $\beta\in[0,1]$. We equip $\calH$ with the inner product
\[
 \big\langle(v,c),(\tv,\tc)\big\rangle_{\mathcal H}
 =\int_0^1 \Big(\mathbb E_{p_\beta}[v_\beta\cdot \tv_\beta]+c_\beta \tc_\beta\Big) \,\mathrm d\beta.
\]

The residual of the log form of the continuity equation is
\[
 R_\beta(v,c)
 =h-c_\beta+\nabla \log p_\beta\cdot v_\beta+\nabla \cdot v_\beta
   =h-c_\beta + p_\beta^{-1}\nabla \cdot(p_\beta v_\beta).
\]
We define the objective $\calF:\calH\to[0,\infty]$ as the squared norm of this residual:
\begin{equation}\label{eq:opt-loss}
 \calF(v,c) =\frac12\int_0^1\EE[p_\beta]{R_\beta(v,c)^2} \,\mathrm d\beta.
\end{equation}
For a scalar or vector field $f$, write $\|f\|_{L^2_\beta(p_\beta)}^2:=\int_0^1\mathbb E_{p_\beta}\|f_\beta\|^2\,\mathrm d\beta$.
Assume that the log density ratio $h\in L^2_\beta(p_\beta)$, and define the domain of $\mathcal F$ by
\[
 \operatorname{dom}\mathcal F
 =\left\{(v,c)\in\mathcal H:
       p_\beta^{-1}\nabla \cdot(p_\beta v_\beta)\in L^2_\beta(p_\beta)\right\}.
\]
The continuity-equation residual loss~\eqref{eq:opt-loss} is also known as a physics-informed neural network (PINN) loss~\citep{mate2023learning,tian2024liouville,chemseddine2025neural,albergo2025nets}. We discuss its practical implementation in Section~\ref{sec:learning-algorithm}.

\subsection{Gradient flow dynamics}

To understand the optimization of $\calF$, we study its idealized gradient flow in the Hilbert space $\calH$. The analysis describes how the residual loss evolves over optimization time and also reveals a connection to the Wasserstein geometry of the annealing path.

For $(v,c)\in\operatorname{dom}\mathcal F$, the Hilbert-space gradient $\nabla_{\mathcal H}\mathcal F(v,c)$ is the unique element of $\mathcal H$ characterized by
\[
\frac{\mathrm d}{\mathrm d\varepsilon}
       \mathcal F(v+\varepsilon u,c+\varepsilon a)\Big|_{\varepsilon=0}
 =\big\langle\nabla_{\mathcal H}\mathcal F(v,c),(u,a) \big\rangle_{\mathcal H},
 \qquad (u,a)\in\operatorname{dom}\mathcal F.
\]
The resulting gradient is
\begin{equation}\label{eq:gradient-formula}
\bigl(\nabla_{\mathcal H}\mathcal F(v,c)\bigr)_\beta=\left(-\nabla R_\beta(v,c),\,-\mathbb E_{p_\beta}R_\beta(v,c)\right).
\end{equation}
Appendix~\ref{sec:derivation-gradient-flow} gives the derivation. The gradient flow of $\mathcal F$, initialized at zero, satisfies
\begin{equation}\label{eq:def-gradient-flow}
 \partial_\tau(v_\tau,c_\tau)
 =-\nabla_{\mathcal H}\mathcal F(v_\tau,c_\tau),
 \qquad (v_0,c_0)=(0,0),
\end{equation}
where $\tau\geq 0$ denotes optimization time, whereas $\beta\in [0,1]$ indexes the annealing path.

\begin{theorem}[Rapid convergence under a uniform Poincar\'e inequality]
\label{thm:pl-inequality}
Suppose that the Poincar\'e inequality~\eqref{eq:component_uniform_poincare} holds for $p_\beta$ with a common constant $C_{\rm P}>0$ over all $\beta\in[0,1]$. Then, for every $(v,c)\in\operatorname{dom}\mathcal F$ with $\nabla R(v,c)\in L^2_\beta(p_\beta)$, the Polyak--\L{}ojasiewicz (PL) inequality holds:
\[
 \|\nabla_{\mathcal H}\mathcal F(v,c)\|_{\mathcal H}^2
 \geq 2 \gamma \mathcal F(v,c),\qquad \gamma=\min\{1,C_{\rm P}^{-1}\}.
\]
Consequently, the loss decays exponentially along the gradient flow of $\calF$:
\[
 \mathcal F(v_\tau,c_\tau)
 \leq e^{-2\gamma\tau}\mathcal F(v_0,c_0),
 \qquad \tau\geq0.
\]
\end{theorem}
The proof is given in Appendix~\ref{prf:pl-inequality}.
Thus, for any $\delta\in(0,1)$, an optimization time of $\frac12\max\{1,C_{\rm P}\}\log(1/\delta)$ suffices to reduce $\calF(v_\tau,c_\tau)$ to at most $\delta$ times its initial value. With zero initialization, this initial value is $\calF(0, 0) = \frac12 \int_0^1 \EE[p_{\beta}]{h^2} \rd\beta $, or one half the squared $L^2$ norm of the log-density ratio $h$ over the annealing path.

Under the componentwise Poincar\'e inequality, Theorem~\ref{thm:componentwise-annealing-length} bounds the componentwise path length. Here, the same condition gives exponential decay of the population residual loss along each component's gradient flow. Thus, when the component paths are sufficiently regular and their endpoints are close, velocity optimization is efficient as well.

Conversely, a long Wasserstein path forces a large loss accumulated over optimization time, as formalized in the following theorem.

\begin{theorem}[Wasserstein length and accumulated loss]
\label{thm:integrated-loss-lower-bound}
Along the gradient flow~\eqref{eq:def-gradient-flow}, the accumulated loss satisfies
\[
 \int_0^\infty\calF(v_\tau,c_\tau)\,\rd\tau
 \geq \frac14\, \calL(p)^2.
\]
\end{theorem}

The integral $\int_0^\infty\calF(v_\tau,c_\tau)\,\rd\tau$ is the area under the loss curve, so the theorem gives a lower bound on the total loss accumulated during optimization. To study the time needed to reach a fixed loss level, we revisit the Gaussian example in Corollary~\ref{coro:gaussian-annealing-length}. For any $\delta>0$, define
\[
 \tau_\delta:=\inf\big\{\tau\geq0:
 \calF(v_\tau,c_\tau)\leq\delta\big\}
\]
as the first time the loss is at most $\delta$.
The following result gives a lower bound, in terms of the dimension $d$, on the time needed to reach this threshold for the Gaussian annealing path.

\begin{proposition}
\label{prop:gaussian-fixed-accuracy}
In the setting of Corollary~\ref{coro:gaussian-annealing-length}, for every fixed $\delta>0$,
\begin{equation*}
 \tau_\delta=\Omega(e^{d/8})
 \qquad\text{as }d\to\infty,
\end{equation*}
where the implicit constant may depend on $\delta$.
\end{proposition}
The proof is given in Appendix~\ref{prf:gaussian-fixed-accuracy}.
Theorem~\ref{thm:pl-inequality} gives loss decay at rate $e^{-2\gamma\tau}$ under a uniform Poincar\'e inequality. For the Gaussian path that transports mass between separated modes, Proposition~\ref{prop:gaussian-fixed-accuracy} instead shows that the time to reach a fixed loss level is at least of order $e^{d/8}$ as $d$ grows. This comparison provides a further optimization motivation for learning flows along localized componentwise paths.

\section{Componentwise annealed transport}
\label{sec:algorithm}

The preceding analysis motivates learning a separate flow along each component path and estimating the component weights independently. We call the resulting sampler \emph{componentwise annealed transport} (CAT). This section describes how to train and sample from the component flows, calibrate their weights, and adapt their annealing schedules.

\subsection{Learning the componentwise flows}
\label{sec:learning-algorithm}

For each component $j\in[J]$, we apply the optimization formulation of Section~\ref{sec:population-residual} to its annealing path from $b_j$ to $p_j$, defined in Section~\ref{sec:mixture_decomposition}. With parameters $\psi\in\Psi$, the functions $v_\psi^j(\beta,x)$ and $c_\psi^j(\beta)$ represent the velocity $v_\beta^j$ and the log-normalizing-constant derivative $c_\beta^j$, respectively. The componentwise analogue of the population objective~\eqref{eq:opt-loss} averages over uniform $\beta\in[0,1]$ and $x\sim p_\beta^j$. In practice, we instead draw $\beta$ from a temperature distribution $\nu^j$ and, conditional on $\beta$, $x$ from a spatial distribution $\mu_\beta^j$. We define the resulting componentwise training loss as
\begin{equation}\label{eq:component-loss}
\begin{aligned}
\mathcal J^j(\psi) &=\E_{\beta\sim\nu^j,X\sim\mu_\beta^j}
  \bigl[R_{\psi}^j(\beta,X)^2\bigr],\\
R_{\psi}^j(\beta,x)
&=h(x)-c_\psi^j(\beta)
  +\nabla\log p_\beta^j(x)
      \cdot v_\psi^j(\beta,x)
  +\nabla \cdot v_\psi^j(\beta,x).
\end{aligned}
\end{equation}
If $\nu^j$ has full support on $[0,1]$ and $\mu_\beta^j$ has full support on the spatial domain for each $\beta$, then $\calJ^j(\psi)=0$ implies $R_\psi^j(\beta,x)=0$ for almost every $(\beta,x)$. Thus, these sampling distributions recover the exact solution at zero loss. Since zero loss is rarely achieved in practice, we choose $\mu^j$ as the distribution path induced by the current velocity field $v_\psi^j$. As the learned velocity field becomes more accurate, the induced path places samples in regions relevant to the intermediate targets. For this choice, $\calJ^j$ also provides an upper bound on the KL divergence between $p_j$ and the learned distribution~\citep[Proposition~5]{albergo2025nets}.

We approximate the population loss $\calJ^j$ with an empirical counterpart. Given collocation pairs $\calD_j=\{(\beta_i^j, X_i^j)\}_{1\leq i\leq N}$ sampled from $\nu^j(\rd\beta)\mu_{\beta}^j(\rd x)$, we define the empirical loss
\begin{align}\label{eq:empirical-loss}
\widehat \calJ^j(\psi;\calD_j) = \frac{1}{|\calD_j|}\sum_{(\beta,x)\in \calD_j } R_{\psi}^j(\beta,x)^2
\end{align}
and minimize it by stochastic gradient descent (SGD). 
For each collocation pair $(\beta,x)$, the score $\nabla\log p_\beta^j(x)=\nabla\log b_j(x)+\beta\nabla h(x)$ is cached and reused.
Algorithm~\ref{alg:component-training} describes the SGD updates with a fixed training set. The updates are \emph{simulation-free}: because the collocation pairs are held fixed, computing the gradient with respect to $\psi$ does not require differentiating through simulated trajectories. By contrast, reverse-KL variational training of a continuous flow typically differentiates through the flow solver. Section~\ref{sec:experiment-training-dynamics} compares the training cost and accuracy of the two approaches.

At each outer iteration, we generate new training samples using the current velocity field. For each collocation pair $(\beta_i^j,X_i^j)$, we draw $\beta_i^j\sim\nu^j$ and obtain $X_i^j=X_{\beta_i^j}^j$ by integrating
\begin{equation}
    \dot X_\beta^j
    =v_\psi^j(\beta,X_\beta^j),
    \qquad \beta\in[0,1],
    \label{eq:particle-ode}
\end{equation}
from $X_0^j\sim b_j$. We add the new pairs to a replay buffer, allowing earlier samples to be reused in subsequent updates. The temperature distribution $\nu^j$ determines where along the path collocation points are collected and thus affects training efficiency. Section~\ref{sec:scheduling} describes how we adapt the corresponding annealing schedule.

\begin{algorithm}[htbp]
\caption{Component-flow training with a fixed training set}
\label{alg:component-training}
\DontPrintSemicolon
\KwIn{fixed training set $\mathcal D_j$, initial parameters $\psi_0$, minibatch size $B$, number of SGD steps $S$, learning rates $(\eta_s)_{s=1}^S$}
\KwOut{updated parameters $\psi_S$}

\For{$s=1$ \KwTo $S$}{
    Draw a minibatch $\mathcal B_s$ of $B$ collocation pairs from $\mathcal D_j$\;
    Compute the minibatch loss $\widehat{\mathcal J}^j(\psi_{s-1};\mathcal B_s)$ using~\eqref{eq:empirical-loss}\;
    Set $\psi_s\gets\psi_{s-1}-\eta_s\nabla_\psi\widehat{\mathcal J}^j(\psi_{s-1};\mathcal B_s)$\;
}
\end{algorithm}

The flow also provides the density of each transported sample. For a fixed velocity field, let $q_\beta^j$ denote the law of $X_\beta^j$ under~\eqref{eq:particle-ode}, with $X_0^j\sim b_j$. Along each trajectory,
\begin{equation*}
    \partial_\beta \log q_\beta^j(X_\beta^j)
    =-\nabla \cdot v_\psi^j(\beta,X_\beta^j),
    \qquad q_0^j=b_j.
\end{equation*}
Write $q^j:=q_1^j$ for the endpoint distribution. Algorithm~\ref{alg:component-sampling} describes how to generate samples and evaluate their densities using the component flows.

\begin{algorithm}[htbp]
\caption{Sampling and density evaluation}
\label{alg:component-sampling}
\DontPrintSemicolon
\KwIn{base density $b_j$, fixed velocity field $v_\psi^j$, sample size $n$}
\KwOut{$n$ independent samples from $q^j$ and corresponding log-densities $\{(X_{i}^j,\ell_{i}^j)\}_{i=1}^n$}

Draw independent $X_{0,i}^j\sim b_j$ and set $\ell_{0,i}^j\gets\log b_j(X_{0,i}^j)$ for $i=1,\ldots,n$\;
For each $i$, solve the coupled ODEs over $\beta\in[0,1]$ to obtain $(X_i^j,\ell_i^j)= (X_{1,i}^j,\ell_{1,i}^j)$:
\[
    \frac{\rd X_{\beta,i}^j}{\rd\beta}
    =v_\psi^j(\beta,X_{\beta,i}^j),
    \qquad
    \frac{\rd\ell_{\beta,i}^j}{\rd\beta}
    =-\nabla\cdot v_\psi^j(\beta,X_{\beta,i}^j).
\]
\end{algorithm}

\subsection{Weight calibration}
\label{sec:weight-calibration}

Recall the target decomposition from Section~\ref{sec:mixture_decomposition}:
\[
    p(x)=\sum_{j=1}^J w_jp_j(x),\qquad
    w_j=\EE[p]{r_j},\qquad
    p_j(x)=\frac{r_j(x)p(x)}{w_j},
\]
where $r_j(x)=a_jb_j(x)/b(x)$ is the responsibility of component $j$. The trained flows approximate $p_j$ by $q^j$; we estimate the corresponding target weights $w_j$ separately.

Write $p=\widetilde p/Z$, where $\widetilde p$ is an evaluable unnormalized target density and $Z=\int\widetilde p(x)\,\rd x>0$ is unknown. The unnormalized mass of component $j$ is $Zw_j=\int r_j(x)\widetilde p(x)\,\rd x$. We estimate this mass by importance sampling from $q^j$: Algorithm~\ref{alg:component-sampling} draws $X_i^j\iid q^j$, $i=1,\ldots,n_j$, and evaluates $q^j(X_i^j)$, giving
\[
    \widehat m^j:=\frac{1}{n_j}\sum_{i=1}^{n_j}W^j(X_i^j),\qquad W^j(x):=\frac{r_j(x)\widetilde p(x)}{q^j(x)}.
\]
Normalizing the mass estimates yields the component weight estimates
\begin{align}\label{eq:hat-w-q}
    \widehat w_j:=\frac{\widehat m^j}{\sum_{\ell=1}^J\widehat m^\ell}.
\end{align}
The following proposition shows that $\widehat w_j$ is consistent for $w_j$ as the sample sizes grow, even when the learned components $q^j$ are inexact.

\begin{proposition}[Consistency of weight calibration]
\label{prop:weight-calibration}
Suppose $q^j>0$ wherever $r_j\widetilde p>0$ for every $j\in[J]$. As $\min_jn_j\to\infty$,
\[  
    \widehat m^j\xrightarrow{\mathrm{a.s.}} Zw_j,
    \qquad
    \widehat w_j\xrightarrow{\mathrm{a.s.}}w_j,
    \qquad j=1,\ldots,J.
\]
The relative variance of $\widehat m^j$ is
\begin{equation}
    \frac{\operatorname{Var}(\widehat m^j)}{Z^2w_j^2}
    =\frac{1}{n_j}\chi^2(p_j\Vert q^j),
    \label{eq:component-weight-variance}
\end{equation}
where $\chi^2(p_j\Vert q^j)=\int p_j(x)^2/q^j(x)\,\rd x-1$ is the $\chi^2$-divergence.
\end{proposition}
The proof is given in Appendix~\ref{prf:weight-calibration}. Consistent weight estimation therefore does not require exact component flows. Equation~\eqref{eq:component-weight-variance} shows that, for a fixed sample size, the relative variance of $\widehat m^j$ is determined by the discrepancy between $q^j$ and its target component $p_j$. If $q^j=p_j$ for every $j$, then every $\widehat m^j$ has zero variance and $\widehat w_j=w_j$ almost surely, regardless of the reference weights $a_j$.

The calibrated approximation $q=\sum_j\widehat w_jq^j$ combines the learned components with their estimated target weights. To estimate expectations, any remaining discrepancy between $q^j$ and $p_j$ can be addressed by componentwise importance sampling. For $f\in L^1(p)$, define the self-normalized importance sampling estimator of $\EE[p_j]{f}$ by
\begin{align*}
    \widehat\mu_j:= \frac{\sum_{i=1}^{n_j}W^j_i f(X^j_i)}{\sum_{i=1}^{n_j}W^j_i}, \qquad W^j_i=W^j(X_i^j).
\end{align*}
Under the assumptions of Proposition~\ref{prop:weight-calibration}, the combined estimator $\sum_j\widehat w_j\widehat\mu_j$ is consistent for $\EE[p]{f}$ as $\min_j n_j\to\infty$.

\subsection{Adaptive annealing schedule}
\label{sec:scheduling}

The difficulty of learning the velocity field $v_\beta^j$ varies with $\beta\in[0,1]$. Intuitively, temperatures at which the field changes rapidly or is highly nonlinear require more training effort. The component-specific temperature measure $\nu^j$ in~\eqref{eq:component-loss} determines how this effort is distributed:
\begin{equation*}
    \mathcal J^j(\psi)
    =\int_0^1
    \E_{X\sim\mu_\beta^j}
    \bigl[R_\psi^j(\beta,X)^2\bigr]\,\nu^j(\rd\beta).
\end{equation*}
In practice, we sample uniformly from a temperature grid in $[0,1]$, corresponding to the discrete measure
\[
    \nu^j_{K_j}=\frac{1}{K_j+1}\sum_{k=0}^{K_j}\delta_{\beta_k^j},\qquad 0=\beta_0^j<\cdots<\beta_{K_j}^j=1.
\]
Thus, regions with a denser grid receive greater weight in the training objective.

We construct the temperature grid adaptively, choosing one increment at a time.
At the current temperature $\beta$, we choose $\Delta$ to control the transport effort from $p_\beta^j$ to $p_{\beta+\Delta}^j$. As discussed in Section~\ref{subsec:wasserstein_geometry}, we measure this effort by the Wasserstein length of the path segment:
\[
    \calL_j(\beta,\Delta)
    :=\int_\beta^{\beta+\Delta}|\dot p^j|(s)\,\rd s,
\]
where $|\dot p^j|(s)$ is the Wasserstein speed of the curve $(p_s^j)_{s\in[0,1]}$ at $s$.
Although $\calL_j(\beta,\Delta)$ is not directly computable, it has a convenient upper bound in terms of the $\chi^2$-divergence between successive targets.

\begin{proposition}
\label{prop:local-annealing-segment-length}
Under the standing regularity conditions and Assumption~\ref{assump:poincare}, for any fixed component \(j\) and temperature \(\beta\in[0,1)\), the following bound holds as \(\Delta\downarrow0\) with \(\beta+\Delta\leq1\):
\begin{equation*}
    \calL_j(\beta,\Delta)^2
    \leq C_{\mathrm P,j}\;
    \chi^2(p_{\beta+\Delta}^j \Vert p_\beta^j)
    +o(\Delta^2).
\end{equation*}
\end{proposition}
The proof is given in Appendix~\ref{prf:local-annealing-segment-length}.
Proposition~\ref{prop:local-annealing-segment-length} motivates choosing the largest increment for which the \(\chi^2\)-divergence between successive targets is at most a prescribed threshold. For sufficiently small increments, the threshold controls the leading term of the upper bound on the squared segment length.

Using the definition of $p_\beta^j$ in~\eqref{eq:local-annealing-path} and the identity $\frac{Z_{\beta+\Delta}^j}{Z_\beta^j} = \E_{p_\beta^j}[e^{\Delta h}]$, we can express the $\chi^2$-divergence as
\begin{align*}
    \chi^2(p_{\beta+\Delta}^j\Vert p_\beta^j)
    &=\E_{p_\beta^j} \left[
        \left(\frac{p_{\beta+\Delta}^j(X)}{p_\beta^j(X)}\right)^2
        \right]-1
    =\frac{\E_{p_\beta^j}[e^{2\Delta h}]}{\bigl(\E_{p_\beta^j}[e^{\Delta h}]\bigr)^2}-1.
\end{align*}
Thus, keeping the $\chi^2$-divergence below a threshold is equivalent to keeping the relative effective sample size (RESS), defined by
\[
    \mathrm{RESS}_j(\beta,\Delta) := \frac{\bigl(\E_{p_\beta^j}[e^{\Delta h}]\bigr)^2}
    {\E_{p_\beta^j}[e^{2\Delta h}]}
    =\frac{1}{1+\chi^2(p_{\beta+\Delta}^j\Vert p_\beta^j)},
\]
above a corresponding threshold. Choosing increments adaptively using relative effective sample size is a common practice in SMC methods~\citep{zhou2016automatic,chopin2020introduction,chopin2024connection}.

We estimate the RESS using the most recently generated samples. At the current temperature $\beta$, let $s=\max\{r\geq0:\bar\beta_r\leq\beta\}$, where $(\bar\beta_r)_r$ are the grid points used to generate these samples and $\bar\beta_0=0$. Let $X_{s,i}^j$, $i=1,\ldots,n$, be independent samples generated at $\bar\beta_s$ with density $q_{\bar\beta_s}^j$. Their generating densities are stored and remain fixed when the flow is updated. Reweight these samples to $p_\beta^j$ and normalize:
\[
    \widetilde\omega_{i}^j
    =\frac{b_j(X_{s,i}^j)e^{\beta h(X_{s,i}^j)}}
    {q_{\bar\beta_s}^j(X_{s,i}^j)},
    \qquad
    \omega_{i}^j
    =\frac{\widetilde\omega_{i}^j}
    {\sum_{\ell=1}^n\widetilde\omega_{\ell}^j}.
\]
The resulting RESS estimate is
\begin{equation*}
    \widehat{N}_j(\beta,\Delta)
    :=\frac{\left(\sum_{i=1}^n
        \omega_{i}^j e^{\Delta h(X_{s,i}^j)}\right)^2}
    {\sum_{i=1}^n\omega_{i}^j e^{2\Delta h(X_{s,i}^j)}}.
\end{equation*}
It equals one at $\Delta=0$ and is nonincreasing for $\Delta\geq0$.
Given a maximum increment $\Delta_{\max}$ and an RESS threshold $N^\star$, choose
\begin{equation}
    \begin{aligned}
        \Delta_j(\beta)
        &=\sup\left\{
            0\leq\Delta\leq\min\{1-\beta,\Delta_{\max}\}:
            \widehat N_j(\beta,\Delta)\geq N^\star
            \right\}.
    \end{aligned}
    \label{eq:transport-schedule-step}
\end{equation}
The next grid point is set to $\beta+\Delta_j(\beta)$.

Algorithm~\ref{alg:cat-adaptive} summarizes CAT with adaptive scheduling.
Because the reference mixture $b$ is fixed, each component can be trained independently and in parallel. For component $j$, we use the same replay buffer $\mathcal D_j$ to train the flow and adapt its schedule, starting with an empty buffer and a temperature grid containing only $0$. In each outer loop iteration, Algorithm~\ref{alg:component-sampling} generates fresh trajectories. We record their states and densities at the current grid points, add the collocation points and target evaluations to $\mathcal D_j$, and train the flow with this data held fixed using Algorithm~\ref{alg:component-training}. We then update the schedule for the next set of trajectories. After training, we calibrate the component weights as described in Section~\ref{sec:weight-calibration}.

\begin{algorithm}[ht]
\caption{Componentwise annealed transport (CAT) with adaptive scheduling}
\label{alg:cat-adaptive}
\DontPrintSemicolon
\KwIn{Unnormalized target $\widetilde p$, reference mixture $b=\sum_j a_jb_j$, target RESS $N^\star\in(0,1)$, maximum increment $\Delta_{\max}>0$, training rounds $M$, SGD steps $S$, minibatch size $B$, learning rates $(\eta_s)_{s=1}^{S}$, trajectory sample size $n$, calibration sample sizes $(n_j)_{j=1}^J$}
\KwOut{Component velocity fields $(v_\psi^j)_{j=1}^J$, weights $(\widehat w_j)_{j=1}^J$, and learned density $q$}

\For{$j=1$ \KwTo $J$ \textnormal{in parallel}}{
    Initialize $v_\psi^j\equiv0$ and $c_\psi^j\equiv0$\;
    Initialize $\mathcal D_j\gets\varnothing$\;
    Set $K_j\gets0$ and $\beta_0^j\gets0$\;
    \For{$m=1$ \KwTo $M$}{
        \tcp{Generate training samples}
        Call Algorithm~\ref{alg:component-sampling} with $b_j$, the current $v_\psi^j$, and sample size $n$; record $(X_{\beta_k^j,i}^j,\ell_{\beta_k^j,i}^j)$ at each grid point $\beta_k^j$\;
        Add these collocation points, their generating proposal densities, and cached values $h(x)$ and $\nabla\log p_\beta^j(x)$ to $\mathcal D_j$\;
        \vspace{0.5em}
        \tcp{Train the velocity field}
        Update $\psi_j$ by calling Algorithm~\ref{alg:component-training} with $\mathcal D_j$, the current $\psi_j$, $B$, $S$, and $(\eta_s)_{s=1}^{S}$\;
        \vspace{0.5em}
        \tcp{Update the schedule}
        Set $\beta_0^j\gets0$ and $k\gets0$\;
        \While{$\beta_k^j<1$}{
            Compute $\Delta_j(\beta_k^j)$ using $\mathcal D_j$ and~\eqref{eq:transport-schedule-step}\;
            Set $\beta_{k+1}^j\gets\beta_k^j+\Delta_j(\beta_k^j)$ and $k\gets k+1$\;
        }
        Set $K_j\gets k$\;
    }
    \vspace{0.5em}
    \tcp{Estimate the component mass}
    Call Algorithm~\ref{alg:component-sampling} with $b_j$, the learned $v_\psi^j$, and sample size $n_j$; retain the endpoint samples $X_i^j$ and densities $q^j(X_i^j)=\exp(\ell_{i}^j)$\;
    Set $\widehat m^j\gets n_j^{-1}\sum_{i=1}^{n_j}r_j(X_i^j)\widetilde p(X_i^j)/q^j(X_i^j)$\;
}
\vspace{0.5em}
\tcp{Weight calibration}
Set $\widehat w_j\gets\widehat m^j/\sum_{\ell=1}^J\widehat m^\ell$ by~\eqref{eq:hat-w-q} and $q\gets\sum_{j=1}^J\widehat w_jq^j$\;
\end{algorithm}

\section{Numerical experiments}\label{sec:experiments}

We evaluate CAT (Algorithm~\ref{alg:cat-adaptive}) on synthetic targets and Bayesian posteriors, summarized in Table~\ref{tab:model-summary}. Section~\ref{sec:experiment-training-dynamics} examines its training dynamics, compares it with related transport methods, and studies how the choice of path affects particle trajectories and mode redistribution. Section~\ref{sec:experiment-comparison} compares CAT with benchmark sampling methods.

\begin{table}[ht]
    \caption{Summary of synthetic targets and models used in the experiments, with the number of observations $n$ (when applicable), and the number of dimensions $d$.}
    \label{tab:model-summary}
    \centering
    \begin{tabular}{lll}
    \toprule
    Distribution/Model & $n$ & $d$ \\
    \midrule
    Skew-normal mixture (Section~\ref{sec:exp-skewgauss2}) & -- & 5 \\
    Skew-$t$ product mixture (Section~\ref{sec:exp-skewt}) & -- & 10 \\
    Seemingly unrelated regression (SUR, Section~\ref{sec:exp-sur}) & 50  & 5 \\
    Input-output hidden Markov model (IOHMM, \citealp{magnusson2025posteriordb}, Section~\ref{sec:exp-iohmm}) & 500 & 29 \\
    Latent Dirichlet allocation (LDA, \citealp{magnusson2025posteriordb}, Section~\ref{sec:exp-lda}) & 4999 & 502 \\
    \bottomrule
    \end{tabular}
\end{table}

All experiments were run on the Duke Compute Cluster using NVIDIA RTX 5000 Ada Generation GPUs, with one GPU allocated to each job. Implementation details are deferred to Appendix~\ref{sec:experiment-supplements}. The code to reproduce the results is available at \href{https://github.com/chenh77/cat.git}{\tt github.com/chenh77/cat}.

\subsection{Training dynamics and transport behavior}\label{sec:experiment-training-dynamics}

We first examine CAT's training dynamics on the IOHMM and LDA posteriors, then compare its computational efficiency with two related transport methods on LDA. Finally, we visualize the learned velocity fields induced by different choices of paths.

\subsubsection{Training dynamics}

\begin{figure}[ht]
    \centering
    \includegraphics[width=.9\linewidth]{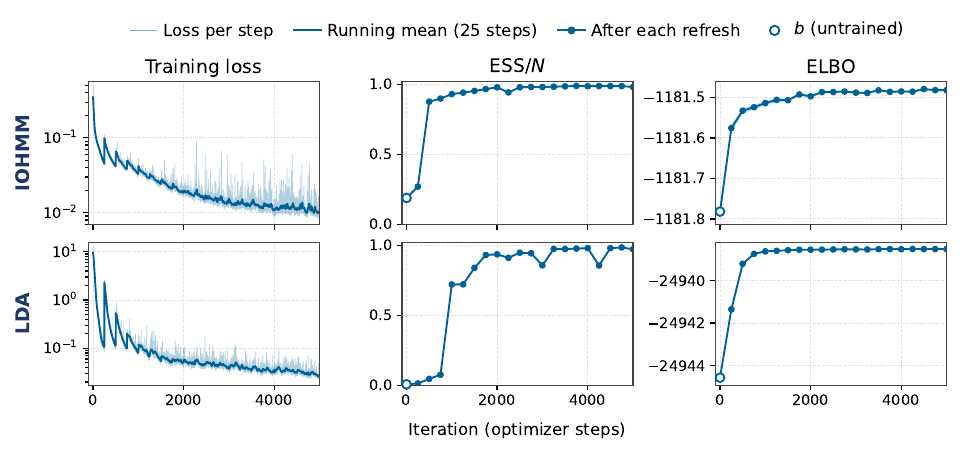}
    \caption{Training dynamics of CAT on IOHMM (top) and LDA (bottom). Columns show the training loss, normalized importance-weight ESS, and ELBO against optimizer steps; ESS and ELBO are computed using calibrated weights. Open circles mark the initial base distribution \(b\) at iteration 0.}
    \label{fig:training-curves}
\end{figure}

We examine CAT's training dynamics on the IOHMM and LDA posteriors. Figure~\ref{fig:training-curves} shows the training loss alongside the normalized importance-weight effective sample size (ESS) and the evidence lower bound (ELBO). As the loss decreases, the ESS approaches one and the ELBO improves, indicating closer agreement between the learned distribution and the target. The loss fluctuates periodically: inner optimization reduces it while the training set is fixed, whereas each outer iteration refreshes the training data and adapts the schedule, which can increase the measured loss. These plots indicate that the training procedure guides the learned distribution toward the target.

\subsubsection{Comparison with related transport methods}

To examine the roles of componentwise paths and prescribed intermediate distributions in CAT, we compare it with two related transport methods:
\begin{itemize}
    \item Globally annealed transport (GAT) starts from the same mixture base $b$ and uses the same continuity-residual training principle as CAT, but follows the global annealing path~\eqref{eq:global_annealing} with a single vector field.
    \item Reverse-KL continuous normalizing flow (RKL-CNF) trains a separate flow for each component, as CAT does, but minimizes the reverse KL divergence to the endpoint component~\citep{rezende2015variational} instead of fitting a prescribed intermediate path through the continuity-residual loss.
\end{itemize}
The comparison with GAT isolates the effect of using componentwise rather than global annealing paths: both methods minimize the continuity-equation residual loss. The comparison with RKL-CNF isolates the effect of fitting prescribed intermediate distributions rather than only the terminal distribution: both methods transport each $b_j$ to $p_j$. CAT and RKL-CNF use the same vector-field parametrization. For GAT, we use a larger single field with approximately as many parameters as CAT's component fields combined; see Appendix~\ref{sec:experiment-vf}. After training CAT and RKL-CNF, we calibrate their component weights as described in Section~\ref{sec:weight-calibration}.

\begin{figure}[ht]
    \centering
    \includegraphics[width=\linewidth]{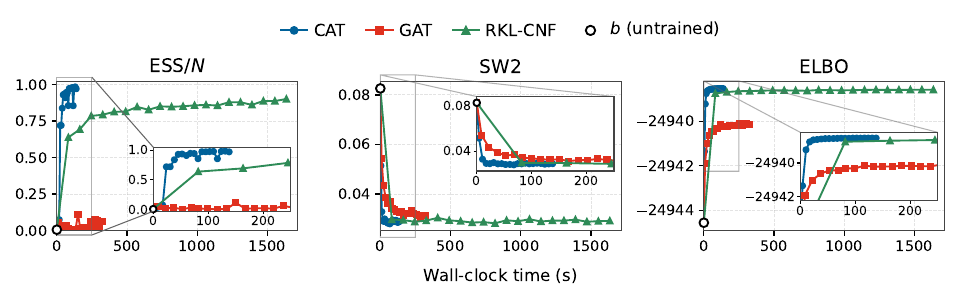}
    \caption{Wall-clock comparison of CAT, GAT, and RKL-CNF on the LDA posterior. Curves show normalized importance-weight ESS, sliced 2-Wasserstein distance (SW2), and ELBO. Open circles give the values for the shared base distribution \(b\) before training; insets enlarge the early training period.}
    \label{fig:lda-wallclock}
\end{figure}

Figure~\ref{fig:lda-wallclock} plots normalized ESS, ELBO, and sliced 2-Wasserstein distance (SW2) against wall-clock time for the LDA posterior. SW2 is computed against reference samples generated using the no-U-turn sampler \citep[NUTS;][]{hoffman2014nuts} from multiple starting points.
CAT reaches a normalized ESS close to one within approximately 100 seconds, whereas RKL-CNF improves more gradually. CAT also attains a comparable ELBO substantially sooner, even though RKL-CNF maximizes the ELBO directly. CAT's shorter wall-clock time stems from its simulation-free training; by contrast, RKL-CNF requires backpropagation through flow simulations, which is time-consuming and memory-intensive. GAT achieves substantially lower ESS and ELBO than the other methods. This result highlights the advantage of separating flow learning from weight calibration.

\subsubsection{Visualization of velocity fields}
\label{sec:experiment-paths}

We visualize the learned velocity fields for CAT and GAT to understand how the choice of annealing path affects the flow. We consider a ten-dimensional distribution which is a mixture of two skew-$t$ components; see Appendix~\ref{sec:exp-skewt} for the definition. A parameter $\gamma$ controls the separation between the two components.
The target has a major mode with weight \(0.8\) and a minor mode with weight \(0.2\), whereas the base distribution $b$ assigns nearly equal weights to the two modes. GAT must therefore redistribute mass between modes through its global flow, whereas CAT transports each component separately and then calibrates the mixture weights.

\begin{figure}[ht]
    \centering
    \includegraphics[width=\linewidth]{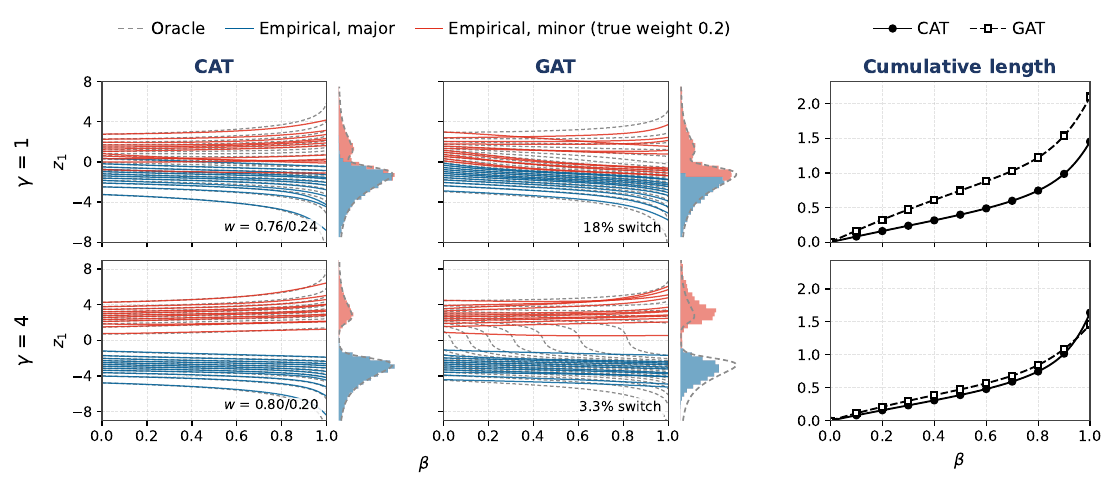}
    \caption{Learned trajectories and cumulative transport lengths for CAT and GAT at \(\gamma=1\) (top) and \(\gamma=4\) (bottom). Left and middle: trajectories projected onto the first coordinate, colored by initial component. Gray dashed curves show oracle trajectories and target marginals; histograms show terminal samples. Annotations give CAT's calibrated mixture weights and GAT's switching rates. Right: cumulative lengths computed from the full ten-dimensional velocity fields.}
    \label{fig:trajectory}
\end{figure}

Figure~\ref{fig:trajectory} compares the learned and oracle trajectories projected onto the first coordinate. CAT closely follows the componentwise oracle at both \(\gamma=1\) (top) and \(\gamma=4\) (bottom). GAT follows the global-path oracle reasonably well at \(\gamma=1\), including its cross-mode transport. At \(\gamma=4\), however, GAT's trajectories remain largely within their initial modes, while the oracle trajectories cross the wider gap. Consequently, GAT assigns incorrect weights to the two modes and produces an overly concentrated marginal within the major mode, underrepresenting its tail. This suggests that learning a single flow to realize both local transport and mass redistribution becomes increasingly difficult as the separation grows.

The right panel of Figure~\ref{fig:trajectory} plots the cumulative transport length \(\int_0^\beta \bigl\{\mathbb E[\|v_t(X_t)\|^2]\bigr\}^{1/2}\,\mathrm{d}t\) against \(\beta\). We estimate the expectation from simulated particles; for CAT, the estimate also averages over components according to their mixture weights. At \(\gamma=1\), GAT has a greater measured length because its flow moves mass between modes. At \(\gamma=4\), the measured lengths are similar because GAT does not realize the required cross-mode redistribution.

\subsection{Comparison with benchmark methods}\label{sec:experiment-comparison}

We compare CAT with benchmark sampling methods on the synthetic two-skew-normal mixture and three Bayesian posteriors: SUR, IOHMM, and LDA. All methods use the same Gaussian mixture base distribution \(b=\sum_j a_j\mathcal N(\mu_j,\Sigma_j)\) constructed from the mode exploration procedure described in Appendix~\ref{sec:experiment-base-training}. We also report results for \(b\) (GMM) itself to assess the improvement over this initial approximation.

The benchmarks are adaptive parallel tempering \citep[APT;][]{miasojedow2013adaptive}, its variant with deterministic even-odd swaps \citep[APT-DEO;][]{okabe2001replica,syed2022nonreversible}, the annealed leap-point sampler \citep[ALPS;][]{tawn2021annealed}, flow annealed importance sampling bootstrap \citep[FAB;][]{midgley2023flow}, TemperFlow \citep[TF;][]{qiu2024efficient}, and annealed flow transport \citep[AFT;][]{arbel2021annealed}. The shared mixture \(b\) provides CAT's component distributions, AFT's annealing reference, the flow base for FAB and TF, and the initial states for APT and APT-DEO. ALPS uses it to construct its tempered targets and leap proposals.
For APT and APT-DEO, we adapt the temperature spacing to target an acceptance rate of \(0.234\) for exchanges between adjacent chains~\citep{atchade2011towards}. Other tuning and scheduling choices are detailed in Appendix~\ref{sec:experiment-base-training}. Reference samples for the Bayesian posteriors are generated using NUTS \citep{hoffman2014nuts} with multiple starting points.

We measure computational cost by the number of target-density evaluations plus \(d\) times the number of target-score evaluations, where \(d\) is the target dimension. We match computational budgets across methods using this measure and repeat each experiment independently 100 times.
\emph{Recovery} is the fraction of runs that visit every mode, and \emph{Mode TV} is the total variation distance between empirical and target mode probabilities (when applicable). We measure distributional accuracy by SW2 to reference samples and report \emph{wall-clock time} for computational efficiency. For methods with an evaluable proposal density \(q\), we also report normalized ESS and \(\operatorname{Var}_{q}(\log w)\), where \(w=p/q\). Table~\ref{tab:summary} reports means and standard deviations for these metrics, except Recovery.

Among methods with evaluable proposal densities, CAT has the highest mean normalized ESS and the lowest mean log-weight variance on all four targets. Its mean sliced Wasserstein distance is also among the lowest for each target. The wall-clock time for CAT is also competitive on the two high-dimensional Bayesian posteriors (IOHMM and LDA).

\begin{table}[ht]
\centering
\small
\setlength{\tabcolsep}{4pt}
\caption{Comparison on four targets over 100 runs. Entries report means \(\pm\) std, except Recovery.  Arrows indicate the preferred direction; dashes denote unreported quantities. }
\label{tab:summary}
\begin{tabular}{lcccccc}
\toprule
Method & Recovery $\uparrow$ & Mode TV $\downarrow$ & SW2 $\downarrow$ & Time (s) & ESS$/N$ $\uparrow$ & $\operatorname{Var}_{q}(\log w)$ $\downarrow$ \\
\midrule
\multicolumn{7}{l}{\textit{Skew-normal mixture}} \\
\midrule
CAT     & \textbf{1.00} & \textbf{\pmsd{0.004}{0.004}} & \textbf{\pmsd{0.254}{0.121}} & \pmsd{61}{5}  & \textbf{\pmsd{0.994}{0.002}} & \textbf{\pmsd{0.01}{0.00}} \\
GMM     & \textbf{1.00} & \pmsd{0.409}{0.009} & \pmsd{2.591}{0.091} & \pmsd{11}{1} & \pmsd{0.536}{0.007} & \pmsd{52.85}{2.31} \\
FAB     & \textbf{1.00} & \pmsd{0.408}{0.008} & \pmsd{2.753}{0.100} & \pmsd{36}{2} & \pmsd{0.559}{0.008} & \pmsd{18.73}{2.96} \\
TF      & \textbf{1.00} & \pmsd{0.407}{0.008} & \pmsd{2.861}{0.099} & \pmsd{37}{2} & \pmsd{0.539}{0.050} & \pmsd{1.23}{0.12} \\
APT     & 0.97 & \pmsd{0.139}{0.109} & \pmsd{1.332}{0.713} & \pmsd{17}{1} & -- & -- \\
APT-DEO & 0.99 & \pmsd{0.109}{0.072} & \pmsd{1.160}{0.527} & \pmsd{13}{1} & -- & -- \\
ALPS    & \textbf{1.00} & \pmsd{0.010}{0.006} & \pmsd{0.257}{0.114} & \pmsd{28}{2} & -- & -- \\
AFT     & \textbf{1.00} & \pmsd{0.007}{0.005} & \pmsd{0.298}{0.123} & \pmsd{20}{2} & -- & -- \\
\midrule
\multicolumn{7}{l}{\textit{SUR}} \\
\midrule
CAT     & \textbf{1.00} & -- & \textbf{\pmsd{0.097}{0.117}} & \pmsd{55}{3} & \textbf{\pmsd{0.717}{0.272}} & \textbf{\pmsd{0.04}{0.02}} \\
GMM     & \textbf{1.00} & -- & \pmsd{0.308}{0.176} & \pmsd{8}{0}  & \pmsd{0.067}{0.056} & \pmsd{20.16}{51.45} \\
FAB     & \textbf{1.00} & -- & \pmsd{1.219}{4.478} & \pmsd{41}{2} & \pmsd{0.020}{0.024} & \pmsd{129.45}{131.97} \\
TF      & \textbf{1.00} & -- & \pmsd{0.445}{0.292} & \pmsd{35}{3} & \pmsd{0.062}{0.053} & \pmsd{3.42}{2.24} \\
APT     & \textbf{1.00} & -- & \pmsd{0.427}{0.195} & \pmsd{15}{1} & -- & -- \\
APT-DEO & \textbf{1.00} & -- & \pmsd{0.419}{0.177} & \pmsd{12}{1} & -- & -- \\
ALPS    & 0.99 & -- & \pmsd{48.740}{305.114} & \pmsd{23}{3} & -- & -- \\
AFT     & \textbf{1.00} & -- & \pmsd{0.115}{0.110} & \pmsd{14}{1} & -- & -- \\
\midrule
\multicolumn{7}{l}{\textit{IOHMM}} \\
\midrule
CAT     & \textbf{0.88} & \textbf{\pmsd{0.042}{0.061}} & \pmsd{0.108}{0.013} & \pmsd{110}{6} & \textbf{\pmsd{0.986}{0.013}} & \textbf{\pmsd{0.01}{0.00}} \\
GMM     & \textbf{0.88} & \textbf{\pmsd{0.042}{0.061}} & \pmsd{0.118}{0.014} & \pmsd{43}{5}  & \pmsd{0.365}{0.138} & \pmsd{0.54}{0.60} \\
FAB     & \textbf{0.88} & \textbf{\pmsd{0.042}{0.061}} & \pmsd{0.111}{0.015} & \pmsd{110}{7} & \pmsd{0.531}{0.113} & \pmsd{0.63}{0.69} \\
TF      & \textbf{0.88} & \textbf{\pmsd{0.042}{0.061}} & \pmsd{0.109}{0.013} & \pmsd{110}{9} & \pmsd{0.519}{0.142} & \pmsd{0.40}{0.07} \\
APT     & 0.00 & \pmsd{0.550}{0.097} & \pmsd{1.058}{0.238} & \pmsd{108}{7} & -- & -- \\
APT-DEO & 0.00 & \pmsd{0.564}{0.105} & \pmsd{1.044}{0.237} & \pmsd{107}{7} & -- & -- \\
ALPS    & \textbf{0.88} & \textbf{\pmsd{0.042}{0.060}} & \textbf{\pmsd{0.107}{0.012}} & \pmsd{117}{6} & -- & -- \\
AFT     & 0.87 & \pmsd{0.105}{0.151} & \pmsd{0.303}{0.431} & \pmsd{63}{6} & -- & -- \\
\midrule
\multicolumn{7}{l}{\textit{LDA}} \\
\midrule
CAT     & \textbf{1.00} & \pmsd{0.006}{0.004} & \textbf{\pmsd{0.029}{0.000}} & \pmsd{243}{12} & \textbf{\pmsd{0.969}{0.048}} & \textbf{\pmsd{0.02}{0.00}} \\
GMM     & \textbf{1.00} & \pmsd{0.006}{0.005} & \pmsd{0.079}{0.002} & \pmsd{44}{4}   & \pmsd{0.003}{0.003} & \pmsd{12.49}{0.28} \\
FAB     & \textbf{1.00} & \pmsd{0.006}{0.004} & \pmsd{0.032}{0.001} & \pmsd{295}{6}  & \pmsd{0.012}{0.008} & \pmsd{12.28}{0.58} \\
TF      & \textbf{1.00} & \textbf{\pmsd{0.005}{0.004}} & \pmsd{0.031}{0.001} & \pmsd{612}{138} & \pmsd{0.017}{0.011} & \pmsd{5.16}{0.47} \\
APT     & 0.96 & \pmsd{0.168}{0.125} & \pmsd{0.035}{0.001} & \pmsd{107}{4} & -- & -- \\
APT-DEO & 0.94 & \pmsd{0.181}{0.136} & \pmsd{0.035}{0.001} & \pmsd{110}{4} & -- & -- \\
ALPS    & \textbf{1.00} & \textbf{\pmsd{0.005}{0.004}} & \textbf{\pmsd{0.029}{0.000}} & \pmsd{3426}{65} & -- & -- \\
AFT     & \textbf{1.00} & \pmsd{0.036}{0.047} & \pmsd{0.032}{0.002} & \pmsd{89}{4} & -- & -- \\
\bottomrule
\end{tabular}
\end{table}

\begin{figure}[ht]
    \centering
    \includegraphics[width=0.9\linewidth]{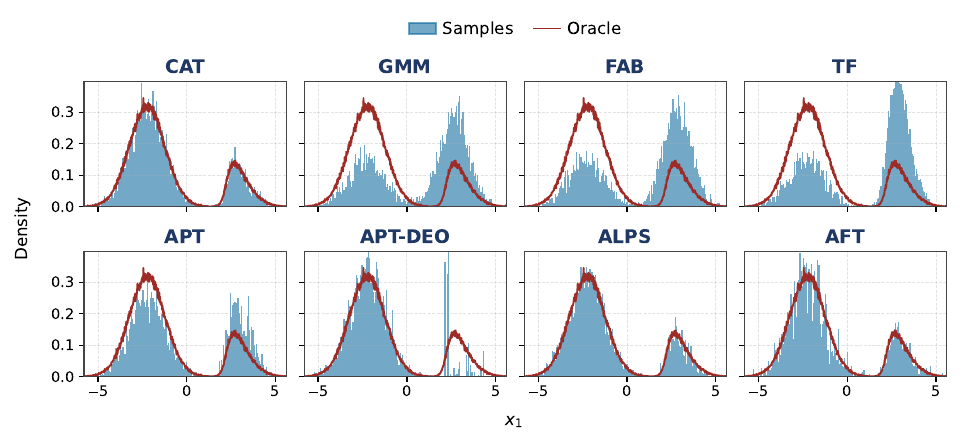}
    \caption{First-coordinate marginals for the skew-normal mixture target. Each panel compares a method's sample histogram (blue) with the exact target marginal density (red). The two modes reveal differences in relative mass and in the shape of the skewed mode.}
    \label{fig:skewgauss2}
\end{figure}

On the skew-normal mixture, CAT substantially improves both the mode weights and the within-mode approximation relative to the initial GMM approximation. FAB and TF leave the GMM's mode-weight error nearly unchanged, whereas CAT, ALPS, and AFT achieve much smaller errors and similar sliced Wasserstein distances. Figure~\ref{fig:skewgauss2} shows that GMM, FAB, and TF assign too much mass to the skewed mode and poorly capture its asymmetric shape. CAT closely matches both the mass and shape of this mode; ALPS and AFT also agree closely with the target marginal.

On SUR, CAT has the lowest mean sliced Wasserstein distance, with AFT close behind. CAT's normalized ESS is substantially higher than those of GMM, FAB, and TF. ALPS has a large and highly variable sliced Wasserstein error, suggesting occasional severe discrepancies despite recovering all modes in 99\% of runs.

On IOHMM, CAT, GMM, FAB, TF, and ALPS have the same recovery rate of \(0.88\), nearly identical mode-weight errors, and similar sliced Wasserstein distances. CAT's clearest advantage lies in the importance-weight diagnostics: its normalized ESS is close to one, and its log-weight variance is much lower. Neither APT nor APT-DEO visits every mode in any run under the allocated budget.

On LDA, most methods recover both modes, but their importance-weight diagnostics differ markedly. GMM, FAB, and TF have small mode-weight errors yet low normalized ESS. CAT combines accurate mode weights with a normalized ESS close to one. CAT and ALPS attain the smallest reported sliced Wasserstein distances, while ALPS takes about 14 times as long as CAT.

\section{Discussion}\label{sec:discussion}

We have investigated the Wasserstein geometry of annealing paths and shown why transport learning is challenging when probability mass moves between well-separated modes. We proposed componentwise annealed transport (CAT) to separate the learning of transport within each component from the estimation of component weights. We also studied how path geometry affects the optimization dynamics. Together, these results highlight the importance of choosing probability paths carefully when designing efficient transport strategies.

The effectiveness of CAT depends on the mixture decomposition of the target $p=\sum_j w_jp_j$ induced by the base distribution $b=\sum_j a_jb_j$. Although this decomposition is exact, its usefulness depends on how readily each $b_j$ can be transported to its corresponding $p_j$. Choosing suitable components may be difficult for targets with overlapping modes, curved high-density regions, or funnel-like structures. One direction is to adapt the decomposition during training by splitting components with difficult internal transport problems and merging those that can be learned effectively together. Diagnostics that guide these decisions and distinguish when a richer base component would be more useful than a split could broaden the applicability of CAT.

Another direction is to reduce the computational cost of training the velocity fields. The continuity equation residual loss \eqref{eq:component-loss} contains a divergence term that can be costly to compute. Weak formulations of the continuity equation offer one potential approach: integration by parts transfers spatial derivatives from the velocity field to test functions~\citep{berman2026leveraging}. The action matching objective~\citep{neklyudov2023action} offers an alternative variational formulation. However, estimating its expectations under the prescribed distribution path using importance weights can lead to high Monte Carlo variance~\cite{albergo2025nets}.

Finally, the learned velocity maps can be incorporated into SMC or parallel tempering to address sampling bias. In SMC, they can guide MCMC rejuvenation steps to improve exploration at each intermediate distribution along the annealing path. Our approach also provides transport maps between any two intermediate distributions along the path. These maps could facilitate faster exchanges between the lowest and highest temperature chains in parallel tempering~\citep{invernizzi2022skipping}, potentially mitigating its random-walk behavior.

\FloatBarrier

\appendix

\section{Proofs}

\subsection{Proof of Lemma~\ref{lem:lower_bound}}
\label{prf:lower_bound}

\begin{proof}[Proof of Lemma~\ref{lem:lower_bound}]
Let \(v_\beta\) be a velocity field realizing the path $(\rho_\beta)_{\beta\in[0,1]}$. The weak form of the continuity equation gives, for almost every $\beta$,
\[
    m_\chi'(\beta)
    =\frac{\rd}{\rd\beta}\int\chi(x)\,\rho_\beta(\rd x)
    =\int\nabla\chi(x)\cdot v_\beta(x)\,\rho_\beta(\rd x).
\]
By Cauchy--Schwarz, \(|m_\chi'(\beta)|\leq s_\chi(\beta)\,\|v_\beta\|_{\rho_\beta}\). 
Taking the infimum over all velocity fields realizing the path gives \(|m_\chi'(\beta)|\leq s_\chi(\beta)\,|\dot \rho|(\beta)\). 
Using the assumption that $s_\chi(\beta)\leq \ep$ for $\beta\in[\beta_0,\beta_1]$, we have
\begin{align*}
    \calL(\rho)&\geq \int_{\beta_0}^{\beta_1} |\dot\rho(\beta)|\,\rd\beta
    \geq \int_{\beta_0}^{\beta_1} \frac{|m_\chi'(\beta)|}{\ep}\,\rd\beta
    \geq \frac{|m_\chi(\beta_1)-m_\chi(\beta_0)|}{\ep}.
\end{align*}
\end{proof}

\subsection{Proof of Corollary~\ref{coro:gaussian-annealing-length}}
\label{prf:gaussian-annealing-length}

\begin{proof}[Proof of Corollary~\ref{coro:gaussian-annealing-length}]
The projection $\mathbf 1_d^\top X/\sqrt d$ has distribution $\mathcal N(-\sqrt d,1)$ under $f_-$ and $\mathcal N(\sqrt d,1)$ under $f_+$. Choose a continuously differentiable function \(\eta:\mathbb R\to[0,1]\) with
\[
    \eta(u)=0\quad(u\leq-1/2),
    \qquad \eta(u)=1\quad(u\geq1/2),
    \qquad \eta(-u)=1-\eta(u),
    \qquad |\eta'(u)|\leq2.
\]
For example, on \([-1/2,1/2]\), take \(\eta(u)=3(u+1/2)^2-2(u+1/2)^3\). Set \(\chi(x)=\eta(\mathbf 1_d^\top x/d)\). Its gradient vanishes outside $B:=\{x:|\mathbf 1_d^\top x|\leq d/2\}$, and we always have $\|\nabla\chi\|\leq2/\sqrt d$.

The geometric annealing density can be written as
\[
    p_\beta(x)
    =\frac{f_-(x)\{1+f_+(x)/f_-(x)\}^{\beta}}{Z_\beta},
    \qquad
    Z_\beta:=\E_{f_-}\!\left[\{1+f_+/f_-\}^{\beta}\right].
\]
For \(0\leq\beta\leq1\), we have \(Z_\beta\geq1\) and \((1+z)^\beta\leq1+z\) for \(z\geq0\). Hence \(p_\beta\leq f_-+f_+\) pointwise. Let \(\widebar\Phi(u)=\bbP(Z\geq u)\) for \(Z\sim\mathcal N(0,1)\). The Gaussian tail bound \(\widebar\Phi(u)\leq\tfrac12 e^{-u^2/2}\), for \(u\geq0\), gives
\[
    \PP[p_\beta]{X\in B}
    \leq2\widebar\Phi(\sqrt d/2)
    \leq e^{-d/8},
    \qquad
    s_\chi^2(\beta)\leq\frac{4}{d}e^{-d/8}.
\]
Symmetry implies \(m_\chi(1)=1/2\), whereas
\[
    m_\chi(0)
    \leq\frac12\widebar\Phi(\sqrt d/2)
        +\frac12\widebar\Phi(\sqrt d)
    \leq\frac12\bigl\{\widebar\Phi(1/2)+\widebar\Phi(1)\bigr\}
    \leq\frac14,
\]
since $\eta(u)\leq\frac12\mathbf 1_{\{u>-1/2\}}+\frac12\mathbf 1_{\{u>0\}}$ and $\widebar\Phi(1/2)+\widebar\Phi(1)<1/2$. Applying Lemma~\ref{lem:lower_bound} over \([0,1]\) yields
\[
    \mathcal L(p)
    \geq\frac{m_\chi(1)-m_\chi(0)}{(2/\sqrt d)e^{-d/16}}
    \geq\frac{\sqrt d}{8}e^{d/16}.
\]
Finally, let \(X\sim f_-\) and let \(U\sim\operatorname{Bernoulli}(1/2)\) be independent of \(X\). Then \(Y=X+2U\mathbf 1_d\) has density \(p\), so
\[
    W_2^2(b,p)\leq\E\|Y-X\|^2=2d.
\]
\end{proof}

\subsection{Proof of Theorem~\ref{thm:componentwise-annealing-length}}
\label{prf:componentwise-annealing-length}

\begin{proof}[Proof of Theorem~\ref{thm:componentwise-annealing-length}]
For almost every \(\beta\), let \(v_\beta^j\) be a minimum-energy velocity for the component path \(p_\beta^j\), so that
\begin{equation*}
    \partial_\beta p_\beta^j+\nabla\cdot(p_\beta^jv_\beta^j)=0,
    \qquad
    |\dot p^j|^2(\beta)=\E_{p_\beta^j}[\|v_\beta^j\|^2].
\end{equation*}
The weighted velocity in~\eqref{eq:def-weighted-velocity} realizes the mixture path because
\begin{align*}
    \partial_\beta \bar p_\beta + \nabla \cdot (\bar p_\beta \bar v_\beta) =
    \sum_{j=1}^J a_j \left[ \partial_\beta p_\beta^j + \nabla\cdot ( p_\beta^j v_\beta^j ) \right]=0.
\end{align*}
Equation~\eqref{eq:wasserstein_speed} and Jensen's inequality then give
\begin{align*}
    |\dot {\bar p}|^2(\beta)
    &\leq \E_{\bar p_\beta}[\|\bar v_\beta\|^2]\leq \sum_{j=1}^J a_j\E_{p_\beta^j}[\|v_\beta^j\|^2]
    =\sum_{j=1}^J a_j|\dot p^j|^2(\beta).
\end{align*}
Thus, by Cauchy--Schwarz on \([0,1]\),
\begin{equation}
    \mathcal L(\bar p)^2
    \leq \int_0^1|\dot {\bar p}|^2(\beta)\,\rd\beta
    \leq \sum_{j=1}^J a_j\int_0^1|\dot p^j|^2(\beta)\,\rd\beta.
    \label{eq:componentwise-action-upper-bound}
\end{equation}

To bound each component speed, note that \(\partial_\beta p_\beta^j=p_\beta^j(h-\E_{p_\beta^j}[h])\). Thus, for any smooth test function \(f\), the continuity equation, Cauchy--Schwarz, and Assumption~\ref{assump:poincare} give
\[
    \E_{p_\beta^j}[\nabla f\cdot v_\beta^j]
    =\operatorname{Cov}_{p_\beta^j}(f,h)
    \leq\bigl\{C_{\mathrm P,j}\operatorname{Var}_{p_\beta^j}(h)\bigr\}^{1/2}
    \bigl\{\E_{p_\beta^j}[\|\nabla f\|^2]\bigr\}^{1/2}.
\]
Taking the supremum over test functions yields 
\[
    |\dot p^j|^2(\beta)
    \leq C_{\mathrm P,j}\operatorname{Var}_{p_\beta^j}(h).
\]
Finally, differentiation along the exponential tilt gives
\(\frac{\rd}{\rd\beta}\E_{p_\beta^j}[h]=\operatorname{Var}_{p_\beta^j}(h)\).
Since \(h=\log(p/b)=\log(p_j/b_j)+\log(w_j/a_j)\), integration over \(\beta\) yields
\begin{align*}
    \int_0^1\operatorname{Var}_{p_\beta^j}(h)\,\rd\beta
    &=\E_{p_j}[h]-\E_{b_j}[h]
    =\mathrm{KL}(p_j\Vert b_j)+\mathrm{KL}(b_j\Vert p_j).
\end{align*}
Combining these identities with~\eqref{eq:componentwise-action-upper-bound} proves the bound.
\end{proof}

\subsection{Derivation of the gradient flow}
\label{sec:derivation-gradient-flow}

Since $R_\beta(v,c)$ is linear in $(v,c)$, for $\nabla R(v,c)\in L^2(p)$ we obtain
\begin{align*}
 \frac{\mathrm d}{\mathrm d\varepsilon}
       \mathcal F(v+\varepsilon u,c+\varepsilon a)\Big|_{\varepsilon=0}
 &=\int_0^1\mathbb E_{p_\beta}\!\left[
 R_\beta(v,c)\left\{p_\beta^{-1}\nabla \cdot(p_\beta u_\beta)-a_\beta\right\}
 \right]\,\mathrm d\beta\\
 &=\int_0^1\!\left\{
 \int_{\mathbb R^d}R_\beta(v,c)\,\nabla \cdot(p_\beta u_\beta)\,\mathrm dx
 -a_\beta\mathbb E_{p_\beta}R_\beta(v,c)
 \right\}\,\mathrm d\beta\\
 &=-\int_0^1\!\left\{\mathbb E_{p_\beta}[\nabla R_\beta(v,c)\cdot u_\beta]
                  +a_\beta\mathbb E_{p_\beta}R_\beta(v,c)\right\}\,\mathrm d\beta,
\end{align*}
where the last equality is obtained by integration by parts.
Comparing this expression with the inner product on $\mathcal H$ gives the gradient formula~\eqref{eq:gradient-formula}.
This gradient $\nabla_{\calH} \calF(v,c)$ belongs to $\calH $ if $(v,c) \in \mathrm{dom} \calF$ and $\nabla R(v,c)\in L^2_\beta(p_\beta)$.

\subsection{Proof of Theorem~\ref{thm:pl-inequality}}
\label{prf:pl-inequality}
\begin{proof}[Proof of Theorem~\ref{thm:pl-inequality}]
By the gradient formula and the Poincar\'e inequality,
\begin{align*}
 \|\nabla_{\mathcal H}\mathcal F(v,c)\|_{\mathcal H}^2
 &=\int_0^1\!\left\{\mathbb E_{p_\beta}\|\nabla_xR_\beta\|^2
                   +(\mathbb E_{p_\beta}R_\beta)^2\right\}\,\mathrm d\beta\\
 &\geq\int_0^1\!\left\{C_{\rm P}^{-1}\operatorname{Var}_{p_\beta}(R_\beta)
                   +(\mathbb E_{p_\beta}R_\beta)^2\right\}\,\mathrm d\beta\\
 &\geq\min\{1,C_{\rm P}^{-1}\}\int_0^1\mathbb E_{p_\beta}[R_\beta^2]\,\mathrm d\beta
 =2\min\{1,C_{\rm P}^{-1}\}\mathcal F(v,c).
\end{align*}
Along the gradient flow, the energy dissipation is 
\[
 \frac{\mathrm d}{\mathrm d\tau}\mathcal F(v_\tau,c_\tau)
 =-\|\nabla_{\mathcal H}\mathcal F(v_\tau,c_\tau)\|_{\mathcal H}^2
 \leq-2\min\{1,C_{\rm P}^{-1}\}\mathcal F(v_\tau,c_\tau).
\]
Gronwall's inequality proves the exponential decay of $\calF(v_\tau, c_\tau)$.
\end{proof}

\subsection{Proof of Theorem~\ref{thm:integrated-loss-lower-bound}}
\begin{proof}[Proof of Theorem~\ref{thm:integrated-loss-lower-bound}]
For almost every $\beta$, let $u_\beta$ be the minimum-norm velocity satisfying the continuity equation, so that $\|u_\beta\|_{L^2(p_\beta)}=|\dot p|(\beta)$.
Since $\partial_\beta p_\beta=p_\beta(h-\EE[p_\beta]{h})$, the continuity equation gives \(p_\beta^{-1}\nabla_x\cdot(p_\beta u_\beta)=\EE[p_\beta]{h}-h\). Therefore,
\begin{align*}
    p_\beta^{-1} \nabla\cdot (p_\beta ( u_\beta - v_{\tau,\beta})) = \EE[p_\beta]{h} - c_{\tau,\beta} - R_\beta(v_\tau,c_\tau).
\end{align*}
By the gradient-flow equation~\eqref{eq:def-gradient-flow},
\(\partial_\tau v_{\tau,\beta}=\nabla R_\beta(v_\tau,c_\tau)\) and
\(\partial_\tau c_{\tau,\beta}=\EE[p_\beta]{R_\beta(v_\tau,c_\tau)}\).
Integration by parts therefore gives
\begin{align*}
\frac{\rd}{\rd\tau}\Bigl\{\|u_\beta-v_{\tau,\beta}\|_{L^2(p_\beta)}^2
 +(\EE[p_\beta]{h}-c_{\tau,\beta})^2\Bigr\}
&=-2\EE[p_\beta]{(u_\beta-v_{\tau,\beta})\cdot\nabla_x R_\beta(v_\tau,c_\tau)}\\
&\quad -2(\EE[p_\beta]{h}-c_{\tau,\beta})\EE[p_\beta]{R_\beta(v_\tau,c_\tau)}\\
&=-2\EE[p_\beta]{R_\beta(v_\tau,c_\tau)^2}.
\end{align*}
The gradient flow converges in $L^2(p_\beta;\mathbb R^d)\times\mathbb R$ to $(u_\beta,\EE[p_\beta]{h})$: the residual is affine in $(v,c)$, and the zero-initialized least-squares gradient flow converges to its minimum-norm zero-residual solution.
Integrating the identity in $\beta$ and $\tau$, then applying Cauchy--Schwarz yields
\begin{align*}
 4\int_0^\infty\calF(v_\tau,c_\tau)\,\rd\tau
 &=\int_0^1\!\left\{\|u_\beta\|_{L^2(p_\beta)}^2+(\EE[p_\beta]{h})^2\right\}\,\rd\beta\\
 &\geq\int_0^1\|u_\beta\|_{L^2(p_\beta)}^2\,\rd\beta
 \geq\left(\int_0^1\|u_\beta\|_{L^2(p_\beta)}\,\rd\beta\right)^2
 =\calL(p)^2.
\end{align*}
\end{proof}

\subsection{Proof of Proposition~\ref{prop:gaussian-fixed-accuracy}}
\label{prf:gaussian-fixed-accuracy}
\begin{proof}[Proof of Proposition~\ref{prop:gaussian-fixed-accuracy}]
Use the cutoff $\chi$ from the proof of Corollary~\ref{coro:gaussian-annealing-length}. That proof gives $\EE[p_1]{\chi}=1/2$, $\EE[f_-]{\chi}\to0$, and, uniformly in $\beta$,
\[
 \|\nabla \chi\|_{L^2(p_\beta)}\leq\frac2{\sqrt d}e^{-d/16}.
\]
Also, $h(x)-h(-x)=2\mathbf1_d^\top x$, so differentiation along the annealing path and reflection give
\[
 \int_0^1\Var[p_\beta]{h}\,\rd\beta
 =\EE[p]{h}-\EE[f_-]{h}
 =\tfrac12\EE[f_+]{h(X)-h(-X)}=d.
\]
Fix $T>0$ and take $d>T$. The density formula in the proof of Corollary~\ref{coro:gaussian-annealing-length}, the bound $(1+a)^\beta\leq1+a^\beta$, and $Z_\beta\geq1$ imply
\[
 \EE[p_{1-T/d}]{\chi}
 \leq\EE[f_-]{\chi}
   +\EE[f_-]{(f_+/f_-)^{1-T/d}}
 =\EE[f_-]{\chi}+e^{-2T+2T^2/d}.
\]
Thus $\limsup_{d\to\infty}\EE[p_{1-T/d}]{\chi}\leq e^{-2T}$. Since $0\leq\chi\leq1$,
\[
 \int_{1-T/d}^1\Var[p_\beta]{\chi}\,\rd\beta\leq\frac{T}{4d}.
\]

Fix $\delta>0$ and suppose first that $\tau_\delta<\infty$. For all sufficiently large $d$, $\calF(0,0)\geq d/2>\delta$. Put $\tau=\tau_\delta$. By continuity of the loss,
\[
 \int_0^1\Var[p_\beta]{R_\beta(v_\tau,c_\tau)}\,\rd\beta
 \leq 2\calF(v_\tau,c_\tau)=2\delta.
\]
Since $\partial_t v_{t,\beta}=\nabla R_\beta(v_t,c_t)$ and the spatial mean of the residual decays at rate one, integration by parts yields
\[
 \frac{\rd}{\rd t}\frac12\int_0^1\Var[p_\beta]{R_\beta(v_t,c_t)}\,\rd\beta
 =-\|\nabla R(v_t,c_t)\|_{L^2_\beta(p_\beta)}^2.
\]
Zero initialization and Cauchy--Schwarz therefore give
\begin{align*}
 \|v_\tau\|_{L^2_\beta(p_\beta)}^2
 &\leq\tau\int_0^\tau\|\nabla R(v_t,c_t)\|_{L^2_\beta(p_\beta)}^2\,\rd t
 =\frac\tau2\left(d-\int_0^1\Var[p_\beta]{R_\beta(v_\tau,c_\tau)}\,\rd\beta\right)
 \leq\frac{\tau d}{2}.
\end{align*}
Differentiating $\EE[p_\beta]{\chi}$ and integrating by parts gives
\[
 \frac{\mathrm d}{\mathrm d\beta}\EE[p_\beta]{\chi}
 =\EE[p_\beta]{(\chi-\EE[p_\beta]{\chi})R_\beta(v_\tau,c_\tau)}
  +\EE[p_\beta]{\nabla_x\chi\cdot v_{\tau,\beta}}.
\]
Integrating over $[1-T/d,1]$ and applying Cauchy--Schwarz yields
\[
 \frac12-\EE[p_{1-T/d}]{\chi}
 \leq\sqrt{\frac{\delta T}{2d}}+\sqrt{2\tau}\,e^{-d/16}.
\]
The same bound holds when $\tau_\delta=\infty$. Hence, for every fixed $T>0$,
\[
 \liminf_{d\to\infty}
 \frac{\tau_\delta}{e^{d/8}}
 \geq\frac12\bigl(\tfrac12-e^{-2T}\bigr)_+^2.
\]
Letting $T\to\infty$ gives
\[
 \liminf_{d\to\infty}\frac{\tau_\delta}{e^{d/8}}
 \geq\frac18>0,
\]
which proves the claimed result.
\end{proof}

\subsection{Proof of Proposition~\ref{prop:weight-calibration}}
\label{prf:weight-calibration}

\begin{proof}[Proof of Proposition~\ref{prop:weight-calibration}]
Since $r_j\widetilde p=Zw_jp_j$, the support assumption gives
\[
    \E_{q^j}[W^j(X)]
    =\int r_j(x)\widetilde p(x)\,\rd x
    =Zw_j<\infty.
\]
Thus $\widehat m^j$ is unbiased for $Zw_j$, and the strong law of large numbers gives $\widehat m^j\to Zw_j$ almost surely. Since $J$ is fixed and $\sum_jw_j=1$, we also have $\sum_j\widehat m^j\to Z>0$ almost surely. The denominator in~\eqref{eq:hat-w-q} is therefore eventually positive, and taking the ratio gives $\widehat w_j\to w_j$ almost surely for every $j$.

The identity $W^j/(Zw_j)=p_j/q^j$ gives
\[
    \frac{\operatorname{Var}_{q^j}(W^j)}{Z^2w_j^2}
    =\int\frac{p_j(x)^2}{q^j(x)}\,\rd x-1
    =\chi^2(p_j\Vert q^j).
\]
Independence within each component gives~\eqref{eq:component-weight-variance}. If $q^j=p_j$ for every $j$, then $W^j=Zw_j$ almost surely, so $\widehat m^j=Zw_j$ and $\widehat w_j=w_j$ for any positive component sample sizes.
\end{proof}

\subsection{Proof of Proposition~\ref{prop:local-annealing-segment-length}}
\label{prf:local-annealing-segment-length}

\begin{proof}[Proof of Proposition~\ref{prop:local-annealing-segment-length}]
Write \(V_j(s)=\operatorname{Var}_{p_s^j}(h)\). The speed bound used in the proof of Theorem~\ref{thm:componentwise-annealing-length} gives
\begin{equation}
    \ell_j(\beta,\Delta)^2
    \leq C_{\mathrm P,j}
    \left(\int_\beta^{\beta+\Delta}\sqrt{V_j(s)}\,\rd s\right)^2.
    \label{eq:segment-length-variance-bound}
\end{equation}
Continuity of \(V_j\) implies that the squared integral in~\eqref{eq:segment-length-variance-bound} is \(V_j(\beta)\Delta^2+o(\Delta^2)\). Also, writing \(M_{\beta,j}(t)=\log\E_{p_\beta^j}[e^{t h}]\), we have
\begin{align*}
    \chi^2(p_{\beta+\Delta}^j\Vert p_\beta^j)
    &=\exp \bigl\{M_{\beta,j}(2\Delta)-2M_{\beta,j}(\Delta)\bigr\}-1\\
    &=V_j(\beta)\Delta^2+o(\Delta^2),
\end{align*}
because \(M_{\beta,j}(0)=0\) and \(M_{\beta,j}''(0)=V_j(\beta)\). This finishes the proof.
\end{proof}

\section{Experiment supplements}\label{sec:experiment-supplements}

This section provides further implementation details of numerical experiments in Section~\ref{sec:experiments}.

\subsection{Base distribution and training budget}\label{sec:experiment-base-training}

For mode exploration, we use the procedure of \citet{tawn2021annealed}, summarized in Algorithm~\ref{alg:mode-explore} for completeness.
The procedure identifies distinct local modes and constructs a Gaussian Laplace anchor around each mode using the local Hessian. These anchors form the base mixture $b$ for subsequent annealing.

\begin{algorithm}[htbp]
\caption{\textsc{ModeExplore}}
\label{alg:mode-explore}
\DontPrintSemicolon
\KwIn{target $p$, exploration inverse-temperature $\beta$, seeds $\{u_i\}_{i=1}^n$, current modes $I=\{(\mu_j,\Sigma_j,a_j)\}_{j=1}^J$, number of hot-chain steps $v$, grouping threshold $c_{\rm M}$}
\KwOut{updated modes $I'=\{(\mu_j,\Sigma_j,a_j)\}_{j=1}^{J'}$, $J'\geq J$}

\For{$i=1$ \KwTo $n$}{
    $u_i'\gets v$ ULA steps targeting $p^\beta$, initialized at $u_i$\;
    Run quasi-Newton optimization initialized at $u_i'$ to obtain a local maximizer $\mu^*$\;
    Compute $\Sigma^*:=-\{\nabla^2\log p(\mu^*)\}^{-1}$ \;
    \If{$J>0$}{
        \For{$j=1$ \KwTo $J$}{
            $d_j\gets d^{-1}
            \max\left\{
                (\mu_j-\mu^*)^\top\Sigma_j^{-1}(\mu_j-\mu^*),
                (\mu_j-\mu^*)^\top(\Sigma^*)^{-1}(\mu_j-\mu^*)
            \right\}$\;
        }
    }
    \If{$J=0$ or $\min_{1\leq j\leq J}d_j>c_{\rm M}$}{
        Append $(\mu^*,\Sigma^*)$ to the mode list and set $J\gets J+1$\;
    }
}
\For{$j=1$ \KwTo $J$}{
    $a_j\gets\displaystyle
    \frac{p(\mu_j)|\Sigma_j|^{1/2}}
    {\sum_{\ell=1}^Jp(\mu_\ell)|\Sigma_\ell|^{1/2}}$\;
}
$J'\gets J$; $I'\gets\{(\mu_j,\Sigma_j,a_j)\}_{j=1}^{J'}$\;
\end{algorithm}

\paragraph{Incorporating modal exploration.}
For the sake of fair comparison, all methods use the same mode information: the modes $\{(\mu_j,\Sigma_j)\}_{j=1}^J$ found by Algorithm~\ref{alg:mode-explore} and the Laplace mixture $b(x)=\sum_{j=1}^J a_j b_j(x)$ built from them, where $b_j=\mathcal N(\mu_j,\Sigma_j)$ and $a_j\propto\widetilde p(\mu_j)|\Sigma_j|^{1/2}$. Wherever a baseline's original design uses an uninformed reference, typically $\mathcal N(0,I)$, we replace it with $b$ and leave the rest of the algorithm unchanged. How $b$ enters therefore follows each method's design:
\begin{itemize}
    \item \emph{Componentwise transport.} CAT and RKL-CNF split the target as $p =\sum_j w_j p_j$ with $p_j\propto a_j b_j\, p/b$, and for each $j$ transport $b_j$ to $p_j$. CAT follows the component-wise annealing path $b_j\exp\{\beta[\log a_j+h]\}$ with $h=\log\widetilde p-\log b$. RKL-CNF has no path and minimizes the reverse KL divergence directly.
    \item \emph{Annealing reference.} GAT and AFT anneal along the geometric path $p_\beta\propto b^{1-\beta} p^{\beta}$, starting from exact draws of $b$.
    \item \emph{Flow base.} FAB and TF use $b$ as the latent base of their normalizing flow. Both flows start as the identity, so training starts from $q=b$. FAB's AIS ladder then runs from the current flow to $p$, as in the original method. TF's path $p^\beta$ has no reference term, so at inverse temperature $\beta$ its base is the tempered mixture $b_\beta=\sum_j a_j(\beta)\,\mathcal N(\mu_j,\Sigma_j/\beta)$ with $a_j(\beta)\propto\widetilde p(\mu_j)^{\beta}|\Sigma_j|^{1/2}$. This is the Laplace approximation of $p^\beta$ at the same modes, and $b_1=b$.
    \item \emph{MCMC.} ALPS uses the mode information as its original design intends: the modes define its weight-preserving tempered targets, a covariance-scaled $b$ is its mode-leaping proposal at the coldest level, and its chains start from scaled versions of $b$. APT and APT-DEO keep their target $p^\beta$ and use $b$ only to initialize the chains.
\end{itemize}
The cost of Algorithm~\ref{alg:mode-explore} is shared and charged to every
method.

\paragraph{Computational cost.}
Let \(N_{\mathrm{val}}\) and \(N_{\mathrm{vg}}\) denote the estimated numbers of target evaluations returning a value and a value and a gradient, respectively. We define the computational budget, subtracting the mutual cost on mode explore, as
\[
    \mathrm{Budget} = N_{\mathrm{val}} + d N_{\mathrm{vg}},
\]
where \(d\) is the target dimension.

\subsection{Learning the vector fields}\label{sec:experiment-vf}

\paragraph{Neural network architecture.}
CAT, GAT and RKL-CNF use the same family of time-conditioned velocity fields. For CAT and RKL-CNF, each discovered mode \(j\), with Gaussian anchor \(b_j=\mathcal N(\mu_j,\Sigma_j)\) and Cholesky factor \(L_j\), has an independent field
\[
    v_j(\beta,x)=L_j f_j(u,\beta),
    \qquad u=L_j^{-1}(x-\mu_j).
\]
GAT uses a single field \(v(\beta,x)=\bar L f(u,\beta)\), with \(u=\bar L^{-1}(x-\bar\mu)\), where \(\bar\mu\) and \(\bar L\bar L^\top\) are the mean and covariance of its Laplace-mixture base \(b=\sum_j a_j b_j\).

The network takes \(u\) concatenated with the time embedding \((\beta,\beta^2,\sin(\pi\beta),\cos(\pi\beta))\). It has three hidden layers of width \(H\), SiLU activations, and a \(d\)-dimensional linear output. Each output coordinate also includes an additive cubic B-spline term \(s_i(\beta,u_i)\), with eight compactly supported basis functions centered uniformly on \([-6,6]\). Its coefficients are generated from the time embedding by a network with one hidden layer of width 32 and SiLU activations. The MLP and spline-coefficient network are trained jointly, with both output layers initialized to zero so that training starts from the identity flow.

The divergence is computed exactly by propagating input Jacobians through the network in a single forward pass, using \(\nabla_x\cdot v_j=\operatorname{tr}(\partial f_j/\partial u)\). We implement Algorithm~\ref{alg:component-sampling} using fixed-step fourth-order Runge--Kutta (RK4), with two steps per grid interval. The widths for CAT and RKL-CNF are listed in Table~\ref{tab:cat-training}; corresponding fields use identical initialization seeds in each run. For GAT, we choose the width whose field parameter count is closest to the combined count of CAT's \(J\) fields. This gives \(H=313,120,163,198\) for the skew-normal mixture, SUR, IOHMM and LDA, respectively.

\begin{table}[ht]
\centering
\caption{CAT training settings. Trajectory counts and minibatch sizes are per mode; minibatches contain collocation points. The width \(H\) is shared by CAT and RKL-CNF.}
\label{tab:cat-training}
\begin{tabular}{lcccc}
\toprule
Target & Refreshes \(\times\) Adam steps & Trajectories & Minibatch & \(H\) \\
\midrule
Skew-normal mixture & \(10\times100\) & 1024 & 256 & 64 \\
SUR                 & \(20\times100\) & 768  & 512 & 64 \\
IOHMM               & \(20\times250\) & 2560 & 512 & 32 \\
LDA                 & \(20\times250\) & 6144 & 512 & 64 \\
\bottomrule
\end{tabular}
\end{table}

All three methods use Adam with a constant learning rate of \(10^{-3}\) and gradient-norm clipping at 5. The optimizer state is retained throughout training.

\paragraph{Training of CAT.}
CAT learns vector fields along the prescribed local annealing path~\eqref{eq:local-annealing-path}. Since the continuity equation does not uniquely determine the velocity field, we penalize its squared norm to favor smaller transport velocities:
\begin{equation*}
    \mathcal J_{\mathrm{reg}}^j(\psi)
    =
    \mathcal J^j(\psi)
    +
    \lambda\,
    \mathbb E_{\beta\sim\nu^j,\,X\sim\mu_\beta^j}
    \bigl[\|v_\psi^j(\beta,X)\|^2\bigr],
\end{equation*}
where \(\mathcal J^j\) is the component log-continuity-equation residual loss defined in~\eqref{eq:component-loss}. We use \(\lambda=10^{-3}\) for all targets. Table~\ref{tab:lambda-sweep} shows that relative ESS changes little over \([10^{-4},3\times10^{-3}]\), supporting this choice as a common default.

\begin{table}[ht]
\centering
\caption{Sensitivity of CAT to the regularization coefficient \(\lambda\). Entries report mean \(\pm\) std of relative ESS over 5 independent runs, separate from those in Section~\ref{sec:experiments}. The default value is shown in bold.}
\label{tab:lambda-sweep}
\begin{tabular}{lcccc}
\toprule
\(\lambda\) & Skew-normal mixture & SUR & IOHMM & LDA \\
\midrule
\(0\)
& \(0.995 \pm 0.004\)
& \(0.763 \pm 0.206\)
& \(0.948 \pm 0.094\)
& \(0.958 \pm 0.038\) \\
\(10^{-4}\)
& \(0.995 \pm 0.004\)
& \(0.797 \pm 0.217\)
& \(0.948 \pm 0.095\)
& \(0.970 \pm 0.012\) \\
\(3\times10^{-4}\)
& \(0.995 \pm 0.004\)
& \(0.786 \pm 0.208\)
& \(0.948 \pm 0.094\)
& \(0.971 \pm 0.014\) \\
\(\boldsymbol{10^{-3}}\)
& \(\boldsymbol{0.995 \pm 0.004}\)
& \(\boldsymbol{0.787 \pm 0.214}\)
& \(\boldsymbol{0.948 \pm 0.094}\)
& \(\boldsymbol{0.969 \pm 0.012}\) \\
\(3\times10^{-3}\)
& \(0.995 \pm 0.004\)
& \(0.788 \pm 0.214\)
& \(0.948 \pm 0.089\)
& \(0.967 \pm 0.013\) \\
\(10^{-2}\)
& \(0.995 \pm 0.003\)
& \(0.786 \pm 0.218\)
& \(0.947 \pm 0.081\)
& \(0.950 \pm 0.046\) \\
\(10^{-1}\)
& \(0.995 \pm 0.002\)
& \(0.853 \pm 0.114\)
& \(0.919 \pm 0.105\)
& \(0.793 \pm 0.195\) \\
\bottomrule
\end{tabular}
\end{table}

The scalar rate \(c_\psi^j\) is trained jointly with the field using a network with two hidden layers of width 32, SiLU activations, the same time embedding, and a zero-initialized output. A fixed mode-specific offset, computed from the first collocation pool, centers the log-density ratio, and its spread sets the scale of the rate-network output. This makes training invariant to additive constants in \(\log\widetilde p\).

Training alternates between trajectory refreshes and stochastic optimization (Algorithm~\ref{alg:cat-adaptive}), with settings in Table~\ref{tab:cat-training}. Each refresh adapts the grid using the latest training or validation trajectories (Section~\ref{sec:scheduling}), then generates fresh trajectories from each anchor under the frozen current field. Their states at grid points form a collocation pool with cached target values and scores. Minibatches sample checkpoints with probabilities determined by the grid spacing, using the trapezoidal rule. Up to 20 recent pools are retained for replay, subject to a memory cap, without additional target evaluations. The target incremental relative ESS is \(0.7\) for the skew-normal mixture and LDA and \(0.3\) for SUR and IOHMM, with \(\Delta_{\max}=0.3\) throughout. For SUR, modes with Laplace weights below \(0.05\) are removed before training.

Every five refreshes and at the end of training, we estimate relative ESS along the current grid using 256 fresh validation trajectories per mode. We return the field and grid from the checkpoint with the highest mean endpoint relative ESS across modes, excluding checkpoints with non-finite validation coordinates or log densities. Validation target evaluations are included in the sampler's reported target-evaluation counts.

\paragraph{Training of GAT.}
GAT transports the Laplace-mixture base \(b=\sum_j a_j b_j\) along the global geometric path~\eqref{eq:global_annealing}. It uses the same regularized residual objective, rate-network architecture, refresh and optimization schedule, replay scheme, validation frequency, and best-ESS checkpoint selection as CAT, applied to a single field. Its own ESS-based grid is constructed from identity-flow draws from \(b\), using the same target relative ESS and maximum increment as CAT. The collocation minibatch size is \(J\) times CAT's per-mode size. We adjust the number of training trajectories so that its oracle cost matches CAT's under the convention in Appendix~\ref{sec:experiment-base-training}.

\paragraph{Training of RKL-CNF.}
RKL-CNF uses the same anchors, field architecture, initialization and RK4 solver as CAT, with a uniform time grid containing the same number of points as CAT's returned adaptive grid. Thus, the two methods use the same number of RK4 steps per trajectory. Each component field is trained by minimizing reverse KL to its endpoint target, using
\[
    \mathcal L_{\mathrm{CNF}}^j(\psi)
    =
    \mathbb E_{X_0\sim b_j}
    \left[
        \log b_j(X_0)
        -\int_0^1 \nabla_x\cdot v_\psi^j(\beta,X_\beta)\,\mathrm d\beta
        -\log\widetilde p_j(X_1)
    \right],
\]
where \(\dot X_\beta=v_\psi^j(\beta,X_\beta)\) and \(\widetilde p_j\) is an unnormalized density of the component endpoint target. This objective equals the reverse KL up to an additive constant independent of \(\psi\). We differentiate through the discretized solver, drawing fresh endpoint training particles at every Adam step. Training uses 20 blocks of 250 steps, with validation every five blocks and after the final block, using 256 fresh trajectories per mode and the same best-ESS selection rule as CAT. Each training particle requires one target value-and-gradient evaluation at the endpoint. We choose the per-mode endpoint batch size so that RKL-CNF's oracle cost matches CAT's under the same budget convention.

\subsection{Further details of examples}\label{sec:experiment-example}

This section describes the target distributions and implementation settings.

\subsubsection{Mixture of two skew normals}\label{sec:exp-skewgauss2}

We construct a five-dimensional mixture of two skew-normal components to illustrate the role of weight calibration (Section~\ref{sec:weight-calibration}):
\[
    \pi(x)
    =0.8\,\mathcal N(x;-c\mathbf 1,I_5)
    +0.2\prod_{j=1}^{5}2\varphi(x_j-c)\Phi\bigl(4(x_j-c)\bigr),
    \qquad c=\sqrt5,
\]
where $\varphi$ and $\Phi$ denote the standard normal density and distribution function. The first component is Gaussian, while the second has independent, positively skewed coordinates. Their location vectors are separated by Euclidean distance 10. The Laplace approximation underestimates the skewed component's mass, providing a test of CAT's weight calibration.

\subsubsection{Mixture of two skew-t distributions}\label{sec:exp-skewt}

We consider the ten-dimensional skew-$t$ product mixture \( \pi(z)=0.8\,\pi_L(z)+0.2\,\pi_R(z) \),
where
\begin{align*}
  \pi_L(z) &=
     \frac{2}{\omega_1}\,t_\nu(y_L)\,
     T_{\nu+1}\left(-\alpha_1 y_L
     \sqrt{\frac{\nu+1}{\nu+y_L^2}}\right)
     \prod_{j=2}^{d}\varphi(z_j),
     & y_L=\frac{z_1+\gamma/2}{\omega_1},\\
  \pi_R(z) &=
     \frac{2}{\omega_1}\,t_\nu(y_R)\,
     T_{\nu+1}\left(\alpha_1 y_R
     \sqrt{\frac{\nu+1}{\nu+y_R^2}}\right)
     \prod_{j=2}^{d}r_c(z_j),
     & y_R=\frac{z_1-\gamma/2}{\omega_1},
\end{align*}
and \(r_c(u)\propto\exp(-u^2/2-cu^4)\) is normalized to integrate to one. Here \(t_\nu\) and \(T_{\nu+1}\) denote the Student-\(t\) density and distribution function with \(\nu\) and \(\nu+1\) degrees of freedom, respectively. The first coordinate thus follows oppositely skewed skew-\(t\) distributions, while the remaining coordinates are Gaussian in the major (left) component and have lighter tails in the minor (right) component. We set \(d=10\), \(\alpha_1=4\), \(\omega_1=2\), \(\nu=3\), and \(c=0.1\), and compare \(\gamma\in\{1,4\}\).

\subsubsection{Seemingly unrelated regression}\label{sec:exp-sur}

The seemingly unrelated regression (SUR) model~\citep{zellner1962efficient, srivastava1987seemingly} comprises multiple regression equations with equation-specific coefficients and correlated errors. Formally, the model consists of $M$ linear regression equations, 
\[
    \bfy_m = X_m \boldsymbol\beta_m + \boldsymbol\varepsilon_m, \qquad m = 1, \dots, M,
\]
where $\bfy_m \in \mathbb{R}^N$ collects the observations for the $m$th equation, $X_m \in \mathbb{R}^{N \times k_m}$ is its design matrix of covariates, and $\boldsymbol\beta_m \in \mathbb{R}^{k_m}$ the corresponding equation-specific regression coefficients. The errors are jointly distributed as
\[
    \boldsymbol{\varepsilon} = (\boldsymbol{\varepsilon}_1, \ldots, \boldsymbol{\varepsilon}_M) \sim \mathcal{N}(0,\, \Sigma \otimes I_N),
\]
with unknown covariance $\Sigma \in \mathbb{R}^{M \times M}$.
Concatenating the coefficients across equations as $\boldsymbol\beta = (\boldsymbol\beta_1, \ldots, \boldsymbol\beta_M)$, inference is typically carried out on the profile likelihood obtained after eliminating $\boldsymbol\Sigma$; this profiling step is what induces multimodality in $\boldsymbol\beta$. We use the example in~\citet{chen2026markov}, which sets $M=5$, $k_1=\cdots=k_M=1$, and yields a target of three well-separated modes.

\subsubsection{Input-output hidden Markov model regression}\label{sec:exp-iohmm}

We consider the input-output hidden Markov model (IOHMM) regression from the \textsf{posteriordb} benchmark suite~\citep{magnusson2025posteriordb}, originally introduced as a \textsf{Stan} case study~\citep{damiano2018hmm}. The data are a simulated series of $T=500$ observations $y_t$ with $M=4$ exogenous inputs $\boldsymbol u_t$, modeled with $K=3$ latent states. Given state $z_t=k$, the output follows a linear Gaussian regression $y_t \sim \mathcal N(\boldsymbol u_t^\top \boldsymbol b_k, \sigma_k^2)$, and the state probabilities depend on the inputs through $\mathrm{softmax}(\boldsymbol u_t^\top \boldsymbol w_1,\dots,\boldsymbol u_t^\top \boldsymbol w_K)$; $\pi_1\in\Delta^{K-1}$ is the initial-state distribution. The priors are 
\[
    \boldsymbol w_k,\boldsymbol b_k \sim \mathcal N(0,5^2 I_M),
    \qquad
    \sigma_k\sim\mathcal N^+(0,3^2),
    \qquad
    \pi_1\sim \mathsf{Unif}(\Delta^{K-1}).
\]  
The latent states are marginalized out with the forward algorithm, and we sample in \textsf{Stan}'s unconstrained space ($\log\sigma_k$ and the simplex transform for $\pi_1$), giving $d=(K-1)+2KM+K=29$. Because no state-identifying constraint is imposed, jointly permuting the labels of $(\pi_{1,k},\boldsymbol w_k,\boldsymbol b_k,\sigma_k)$ leaves the posterior invariant, so it has an exact $K!=6$-fold label-switching symmetry with equal mode weights. In addition, the state probabilities are invariant to adding a common vector to all $\boldsymbol w_k$, so $\boldsymbol w$ is only weakly identified and its posterior is largely prior-dominated.

\subsubsection{Latent Dirichlet allocation}\label{sec:exp-lda}

We consider a latent Dirichlet allocation (LDA) model~\citep{blei2003latent} with $K=2$ topics from the \textsf{posteriordb} benchmark suite~\citep{magnusson2025posteriordb}.
The corpus contains $M=6$ documents, $V=249$ vocabulary terms, and $N=4999$ tokens. Each document $m$ has a topic distribution $\boldsymbol\theta_m$, and each topic $k$ has a word distribution $\boldsymbol\phi_k$. Marginalizing the per-token topic indicators gives the likelihood
\[
    p(\boldsymbol x\mid\boldsymbol\theta,\boldsymbol\phi)
    =
    \prod_{n=1}^{N}
    \sum_{k=1}^{K}
    \theta_{d(n),k}\,\phi_{k,x_n},
\]
where $x_n$ is the $n$th observed token and $d(n)$ denotes its document. We assign symmetric Dirichlet priors,
\[
    \boldsymbol\theta_m\sim\mathrm{Dirichlet}(\mathbf 1_K),
    \quad m=1,\dots,M,
    \qquad
    \boldsymbol\phi_k\sim\mathrm{Dirichlet}(\mathbf 1_V),
    \quad k=1,\dots,K.
\]
Jointly swapping the two topic labels in all document and word distributions leaves the posterior invariant, yielding two equivalent modes related by an exact label-switching symmetry.

Sampling uses an unconstrained parameterization of dimension $d=M(K-1)+K(V-1)=502$. Since the default simplex parameterization induces appreciable skewness in the posterior components, we use a prior-Gaussianizing parameterization that combines the Dirichlet stick-breaking representation~\citep{connor1969concepts} with univariate probability integral transforms. This gives more symmetric, approximately Gaussian local posterior geometry in this example.

For a $D$-dimensional simplex with prior $\boldsymbol x\sim\mathrm{Dirichlet}(\mathbf 1_D)$, let $z_j\stackrel{\mathrm{iid}}{\sim}\mathcal N(0,1)$ and define
\[
    u_j=\Phi(z_j),
    \qquad
    v_j
    =
    F^{-1}_{\mathrm{Beta}(1,D-j)}(u_j)
    =
    1-(1-u_j)^{1/(D-j)},
    \qquad j=1,\dots,D-1.
\]
The simplex coordinates follow by stick breaking:
\[
    x_j=v_j\prod_{\ell<j}(1-v_\ell),
    \quad j=1,\dots,D-1,
    \qquad
    x_D=\prod_{\ell=1}^{D-1}(1-v_\ell).
\]
This bijection maps $\mathcal N(0,I_{D-1})$ to $\mathrm{Dirichlet}(\mathbf 1_D)$ and has a closed-form Jacobian. Applying it to every $\boldsymbol\theta_m$ and $\boldsymbol\phi_k$ makes the prior exactly standard Gaussian in the inference coordinates while preserving the prior and posterior on the original simplex parameters. Samples are mapped back to simplex space for evaluation; posterior-comparison metrics use reference draws transformed into the same $502$-dimensional inference coordinates.

\bibliographystyle{apalike}
\bibliography{ref}

\end{document}